\documentclass{article}

\PassOptionsToPackage{numbers,sort&compress}{natbib}
\usepackage[preprint]{neurips_2026}
\usepackage[utf8]{inputenc}
\usepackage[T1]{fontenc}
\usepackage{hyperref}
\usepackage{url}
\usepackage{booktabs}
\usepackage{amsfonts}
\usepackage{nicefrac}
\usepackage{microtype}
\usepackage{xcolor}
\usepackage{colortbl}
\usepackage{graphicx}
\usepackage{multirow}
\usepackage{array}
\usepackage{makecell}
\usepackage{tabularx}
\usepackage{caption}
\usepackage{subcaption}
\usepackage{listings}
\usepackage{enumitem}
\usepackage{amsmath}
\usepackage{amssymb}
\usepackage[numbers,sort&compress]{natbib}
\usepackage{xspace}
\newcommand{\bench}{\textsc{WildRoadBench}\xspace}
\newcommand{\modeA}{\textsc{VLM\,Track}\xspace}
\newcommand{\modeB}{\textsc{Agent\,Track}\xspace}
\newcommand{\best}[1]{\textbf{#1}}
\newcommand{\second}[1]{\underline{#1}}

\usepackage[most]{tcolorbox}
\usepackage{xcolor}
\usepackage{enumitem}
\usepackage{adjustbox}
\definecolor{stageblue}{HTML}{2563EB}
\definecolor{stagegreen}{HTML}{059669}
\definecolor{stageorange}{HTML}{EA580C}
\definecolor{stagepurple}{HTML}{7C3AED}
\definecolor{stagered}{HTML}{DC2626}
\definecolor{stagecyan}{HTML}{0891B2}
\definecolor{stagepink}{HTML}{DB2777}
\definecolor{stagegrayframe}{HTML}{475569}
\definecolor{stagebg}{HTML}{F8FAFC}

\newtcolorbox{stagebox}[3][]{
  enhanced,
  breakable,
  colback=stagebg,
  colframe=#1,
  boxrule=0.5pt,
  arc=0.8mm,
  boxsep=0.3mm,
  left=0.55em,
  right=0.55em,
  top=0.45em,
  bottom=0.35em,
  title=\textbf{#2 -- #3},
  coltitle=white,
  colbacktitle=#1,
  fonttitle=\small,
  fontupper=\small,
  attach boxed title to top left={
    xshift=0.6em,
    yshift=-1.2mm
  },
  boxed title style={
    boxrule=0pt,
    sharp corners,
    left=0.45em,
    right=0.45em,
    top=0.12em,
    bottom=0.12em
  },
  before skip=0.45em,
  after skip=0.45em
}

\title{\bench: A Wild Aerial Road-Damage \\ Grounding Benchmark for
Vision--Language Models and Autonomous Agents}

\author{%
Bingnan Liu\textsuperscript{1,*} \quad
Chenhang Cui\textsuperscript{2,*} \quad
Rui Huang\textsuperscript{1,*} \quad
Jiani Luo\textsuperscript{2} \quad
Zhirong Shen\textsuperscript{1}
\\[0.4em]
\bfseries
Tinghao Wang\textsuperscript{1} \quad
Xiande Huang\textsuperscript{3} \quad
Lingbei Meng\textsuperscript{4} \quad
Fei Shen\textsuperscript{2} \quad
An Zhang\textsuperscript{5}
\\[0.9em]
\normalfont\small
\textsuperscript{1}University of Electronic Science and Technology of China \quad
\textsuperscript{2}National University of Singapore
\\
\normalfont\small
\textsuperscript{3}De Artificial Intelligence Lab \quad
\textsuperscript{4}The Chinese University of Hong Kong, Shenzhen
\\
\normalfont\small
\textsuperscript{5}University of Science and Technology of China
\\[0.4em]
\normalfont\footnotesize\textsuperscript{*}Equal contribution.}
\begin{document}
\maketitle

\begin{abstract}
We introduce \bench, a wild aerial road-damage grounding benchmark
that couples direct visual grounding by vision--language models with
autonomous research-and-engineering by LLM-driven agents on a single
professionally annotated UAV corpus.  The same image set and the
same per-class AP$_{50}$ metric are evaluated under two protocols.
The \modeA{} measures whether a fixed VLM can localise
domain-specific damage from one image and one short prompt under a
unified prompting, decoding and parsing pipeline.  The \modeB{}
measures whether an autonomous agent, given only a written task
brief, a small exploratory slice and a fixed interaction budget, can
search the public web, adapt pretrained components, write training
and inference code, and submit predictions through a scalar-feedback
oracle on a hidden holdout.  We benchmark a broad pool of 
close-sourced frontier models and open-source VLMs together with several frontier
LLM-driven agents.  Both routes remain far from reliable performance
in this wild setting: close-sourced frontier models lead the VLM
leaderboard but still leave more than half of the metric on the
table; open-source grounders plateau well below them, and newer
generations or reasoning-style variants do not consistently improve
grounding; small targets collapse for every open-source model;
agents lag the strongest VLM despite richer affordances, and
several fail to land a valid submission within the budget.  We
release the code and data in \url{https://github.com/liangyunuancha/wildroadbench} to support reproducible follow-up
research.
\end{abstract}

\section{Introduction}
\label{sec:intro}

Visual grounding is becoming a central test of whether multimodal
systems can move from language-level understanding to physical-world
action: an embodied agent, inspection platform or maintenance assistant
must not only describe an image, but also localise the object, defect or
region that requires intervention.  Recent progress in large
vision--language models (LVLMs) such as GPT-5~\cite{openai2025gpt5mini}, Claude~\cite{anthropic2025claude}, Qwen3.5~\cite{qwen3vl2025} and
InternVL3~\cite{internvl2024} has made this goal appear increasingly attainable; in
parallel, LLM-driven agents have begun to write code, use tools, search
the web and assemble task-specific pipelines on software-engineering,
web-navigation and data-analysis tasks.

This raises two related questions about grounding.  First, if a system
is used as a fixed VLM, can it localise the relevant regions directly
from an image and a text prompt?  Second, if the same localisation
problem is presented as an autonomous research-and-engineering task,
can an LLM-driven agent use examples, tools, external resources and
metric feedback to build a working training and inference pipeline
under a fixed budget?  In both cases the required output is the same
kind of object-level prediction---bounding boxes with confidence
scores---but the route to producing those predictions is different:
one relies on the pretrained VLM's existing grounding ability, while
the other relies on the agent's ability to construct a detector for the
task.

We instantiate these two questions in \textbf{wild aerial road-damage
grounding}.  The task is built from UAV imagery used for
road-maintenance inspection \citep{zhu2022pavement,chen2024pavement,feitosa2024pavement}, with damage regions annotated by
road-maintenance experts following highway-maintenance specifications.
The setting contains small, visually subtle and domain-specific defects
that may be confused with shadows, water stains, seams, lane markings
or ordinary asphalt texture.  It therefore tests whether current
systems can localise professional inspection concepts in noisy
real-world aerial imagery, either through a fixed VLM query or through
an autonomous detector-building workflow.

\bench packages this setting as a benchmark with two evaluation
settings on the same 1{,}061-image corpus and the same 
AP$_{50}$ scoring \citep{everingham2010pascal} (Section~\ref{sec:eval}).  In the
\modeA{}, a VLM receives one image and one target-scene prompt,
performs a single forward pass, and returns a list of bounding boxes
with confidence scores.  This isolates the localisation ability of a
fixed multimodal model without examples, tools or task-specific
training.  In the \modeB{}, an LLM-driven agent receives a written task
document, the target label schema and a small set of annotated samples;
inside an isolated sandbox it may search the public web, inspect
external datasets, download pretrained detectors, install
dependencies, complete training and inference code,  and
submit predictions.  The grader returns only a top-line mAP$_{50}$
score, with limited time
budget.

Across twenty-five VLMs (sixteen open-source releases plus nine
close-sourced frontier models) and 15 frontier LLM-driven agents, both evaluation
settings reveal substantial gaps on this wild localisation problem.
The strongest VLM is Gemini\,3\,Pro~\cite{google2025gemini3} at \textbf{42.1\,\% macro
AP$_{50}$}; the strongest {open-source} VLM,
Qwen2.5-VL-32B-Instruct-AWQ~\cite{qwen25vl2024}, reaches only \textbf{25.5\,\% macro
AP$_{50}$}; the strongest autonomous agent, Claude Opus 4.7 achieves the strongest agent-track result (0.1676) after five attempts in five hours.
Model scaling and newer generations do not consistently translate into
better localisation---some newer VLMs produce many more false
positives, and reasoning-style variants can perform worse than their
non-reasoning counterparts---and small targets remain especially
challenging.  For agents, the benchmark further exposes the difficulty
of turning task descriptions, external resources and sparse metric
feedback into a working aerial road-damage detector.

This paper makes three contributions.  \textbf{First}, we introduce
\bench, a wild aerial road-damage grounding benchmark with
professionally annotated ground truth, built from a 1{,}061-image UAV
corpus.  \textbf{Second}, we define a unified two-track evaluation protocol that
studies grounding from two complementary perspectives: direct VLM
grounding, and automated research-and-engineering by LLM-driven agents
on the same perception task. Both tracks use the same image set and
evaluation metric, while the agent track is implemented through a
controlled harness that manages the workspace, tool use, interaction
budget, submissions, and evaluation.   \textbf{Third}, we benchmark 25 VLMs (16 open-source + 9
close-sourced models) and 15 frontier LLM-driven agents,  and show that wild grounding remains
challenging for both multimodal perception and autonomous agent
workflows.

\begin{figure}[!htbp]
  \centering
  \includegraphics[width=0.9\linewidth]{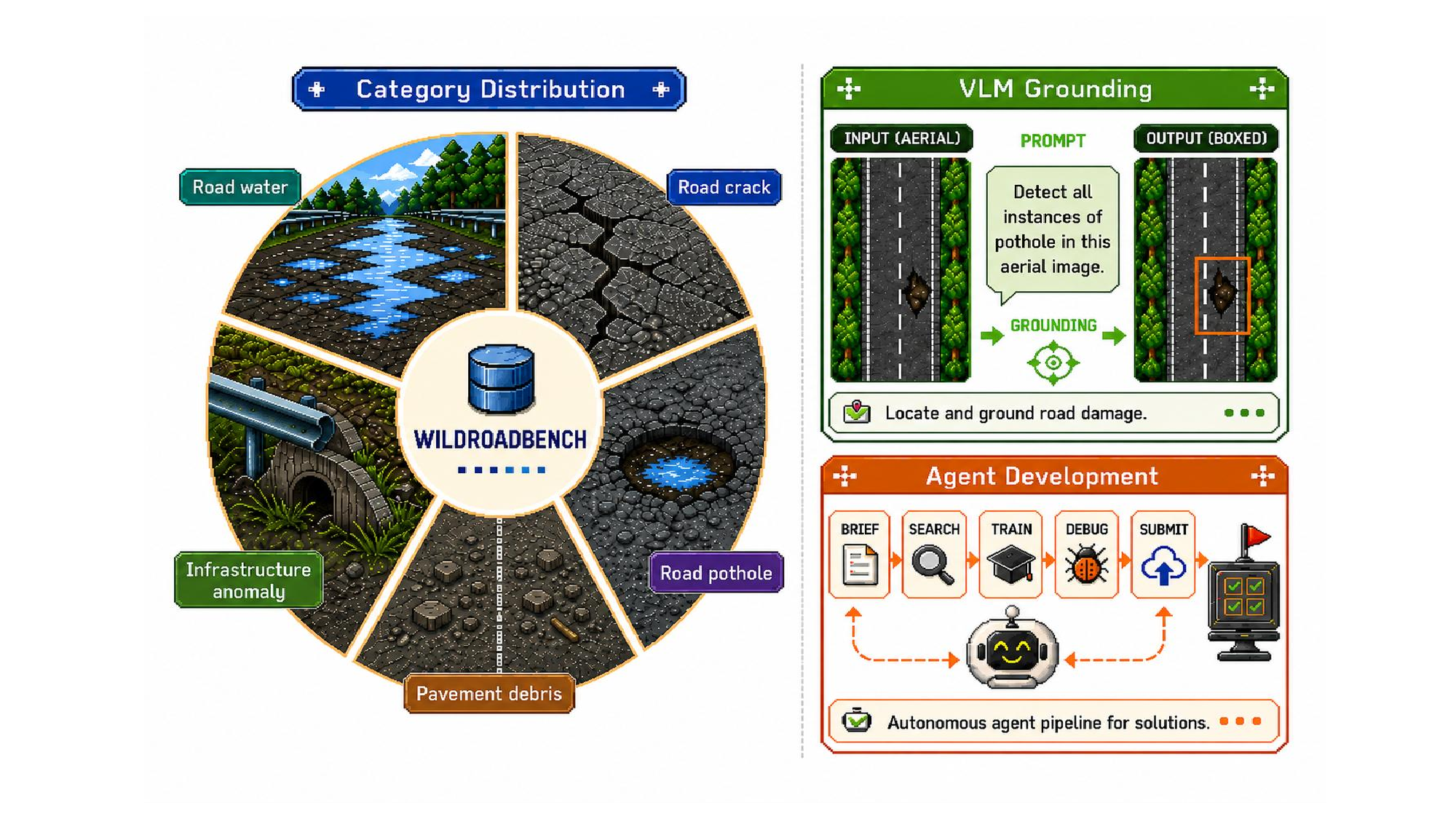}
  \caption{\textbf{\bench at a glance.}  A single 1{,}061-image
  corpus of professionally-annotated UAV road-damage imagery is
  graded under two protocols: the \modeA{} evaluates grounding
  VLMs in a single forward pass against a schema (released
  with images and labels), while the \modeB{} evaluates autonomous
  LLM-driven agents against the underlying schema as a hidden holdout; agents may browse the web, train
  detectors and submit up to five times within a 5\,h wall-clock.
  The grader returns COCO-style AP$_{50}$.}
  \label{fig:overview}
\end{figure}

\section{Related Work}
\label{sec:related}

\paragraph{Grounding and road-damage benchmarks.}
RefCOCO/+/g~\cite{kazemzadeh2014refer,yu2016modeling},
Visual~Genome~\cite{krishna2017visual} and recent spatial probes
such as DORI~\cite{li2025dori}, SSI-Bench~\cite{wang2025ssi} and
MMSI-Bench~\cite{chen2025mmsi} evaluate grounding on
ground-level or internet imagery, where reported performance is
often high but the setting differs substantially from aerial road
inspection.  Road-damage datasets such as
RDD2022~\cite{arya2022rdd2022} focus on ground-level smartphone
imagery, while aerial datasets such as VisDrone~\cite{zhu2018visdrone}
cover pedestrians and vehicles rather than road-surface defects. Together, these benchmarks provide strong coverage of grounding,
road damage, and aerial perception, but they do not jointly evaluate
grounding on aerial road-surface defects or compare VLMs and
LLM-driven agents on the same localisation task.

\paragraph{Autonomous LLM agents on perception tasks.}
Recent agent frameworks such as SWE-Agent~\cite{yang2024sweagent}, OpenHands,
nano-claude-code~\cite{nanocc2024} and CheetahClaws show strong results on symbolic
benchmarks including SWE-Bench~\cite{jimenez2024swebench}, WebArena~\cite{zhou2023webarena} and GAIA~\cite{mialon2023gaia}. 
Recent interactive-agent benchmarks further emphasize tool use and dynamic user--environment interaction in real-world settings, such as VitaBench~\cite{he2025vitabench} and $\tau^2$-Bench~\cite{barres2025tau2bench}.
Automated machine-learning benchmarks~\cite{chan2024mlebench} share the goal of evaluating agents as
end-to-end problem solvers, but typically target tabular or Kaggle-style
tasks.  In contrast, \bench{} evaluates LLM-driven agents on a
visual grounding task with a standard detection metric, testing
whether tool use, data acquisition and code generation can produce a
working detector for the same problem used in
the VLM setting.

\section{The \bench{} Dataset}
\label{sec:data}

\subsection{Source and Collection}
\bench{} is built from internally collected UAV imagery of road
damage on Chinese roads. The images were captured by quad-rotor
drones at 50--200~m altitude. Professional annotation was
conducted by road-maintenance experts. Annotators provided both bounding
boxes and damage categories. The final \textbf{1{,}061}-image \modeA{}
evaluation set contains \textbf{1{,}699} ground-truth boxes spanning
five scene-level categories, as summarised in Table~\ref{tab:scene_dist}.

\providecommand{\tabheader}{}\definecolor{tabheader}{HTML}{1E3A8A}
\providecommand{\tabheadertx}{}\definecolor{tabheadertx}{HTML}{FFFFFF}
\providecommand{\tabopenwin}{}\definecolor{tabopenwin}{HTML}{ECFDF5}
\providecommand{\tabstripe}{}\definecolor{tabstripe}{HTML}{F8FAFC}
\providecommand{\tabrule}{}\definecolor{tabrule}{HTML}{CBD5E1}
\begin{table}[!htbp]
  \centering
  \caption{\textbf{\modeA{} and \modeB{} scene schema on the shared 1{,}061-image corpus.}
  Long-tail infrastructure defects are grouped as
  \textit{infrastructure anomaly}. ``Avg/img'' is the mean GT boxes per
  positive image. See Appendix~\ref{app:original_scene_dist} for more details.}
  \label{tab:scene_dist}

  \scriptsize
  \setlength{\tabcolsep}{7pt}
  \renewcommand{\arraystretch}{1.12}
  \arrayrulecolor{tabrule}

  \begin{adjustbox}{max width=0.45\linewidth}
  \begin{tabular}{l r r r}
    \arrayrulecolor{tabheader}\specialrule{1pt}{0pt}{0pt}
    \rowcolor{tabheader}
    \color{tabheadertx}\textbf{Scene} &
    \color{tabheadertx}\textbf{Images} &
    \color{tabheadertx}\textbf{GT boxes} &
    \color{tabheadertx}\textbf{Avg/img} \\
    \arrayrulecolor{tabheader}\specialrule{1pt}{0pt}{0pt}
    \arrayrulecolor{tabrule}

    \rowcolor{tabstripe}
    Road water             & 624 & 786 & 1.3 \\
    Road crack             & 402 & 561 & 1.4 \\
    \rowcolor{tabstripe}
    Road pothole           & 145 & 175 & 1.2 \\
    Pavement debris        &  48 &  85 & 1.8 \\
    \rowcolor{tabstripe}
    Infrastructure anomaly &  83 &  92 & 1.1 \\

    \arrayrulecolor{tabrule}\specialrule{0.6pt}{1pt}{1pt}
    \rowcolor{tabopenwin}
    \textbf{Total} &
    \textbf{1{,}061$^{\dagger}$} &
    \textbf{1{,}699} &
    \textbf{1.3} \\

    \arrayrulecolor{tabheader}\specialrule{1pt}{0pt}{0pt}
    \arrayrulecolor{black}
  \end{tabular}
  \end{adjustbox}
\end{table}

\subsection{Shared Corpus and Evaluation Settings}
\label{sec:corpus}

\bench{} is built on a single professionally annotated corpus of
1{,}061 road-scene images.  This shared corpus supports two
evaluation settings that target different capabilities under a
common visual domain and a common COCO-style AP$_{50}$ evaluation
rule.  The first setting evaluates direct visual grounding by
fixed VLMs, while the second evaluates whether an autonomous
LLM-driven agent can assemble a working detector through data
acquisition, model adaptation, training and inference.
The two settings differ in the information exposed to the
evaluated system and in the form of interaction with the
benchmark.  In the VLM setting, the images and annotations are
released for offline evaluation, so models can be queried
directly and their predictions can be analysed at the
per-instance level.  In the Agent setting, the same image pool is
used as a hidden holdout: the agent receives only a written task
brief, a small set of exposed examples and scalar submission
feedback, and must return predictions through the prescribed
evaluation interface.

\subsubsection{VLM Setting: Offline Grounding Evaluation}
\label{sec:vlm_setting}

The VLM setting measures direct zero-shot visual grounding.  Each
model is given an image and a short natural-language prompt that
names the target road-scene category, and it is asked to return
axis-aligned bounding boxes with confidence scores. This protocol is
consistent with recent grounded VLM and open-vocabulary detection
settings\cite{liu2023groundingdino,peng2023kosmos2,bai2023qwenvl,you2023ferret}. Because the
evaluation is fully offline, the images, ground-truth boxes and
class labels are released together with the benchmark, enabling
reproducible per-prediction analysis without a submission server.

This setting is designed to evaluate the localisation ability of
a fixed VLM rather than its ability to acquire data or train
task-specific models.  All models are therefore evaluated under a
unified prompting, decoding and parsing protocol, with
family-specific output normalisation applied only to reconcile
different bounding-box coordinate conventions.

\subsubsection{Agent Setting: End-to-End Autonomous Detector Development}
\label{sec:agent_setting}

The Agent setting evaluates whether an LLM-driven agent can complete the
full detector-development loop under a controlled harness. Given a task
specification, a small set of exposed examples, and a fixed interaction
budget, the agent must plan experiments, use tools, search for external
data, adapt pretrained components, write training and inference code, and
submit predictions for hidden-holdout evaluation. The holdout images and
labels are never exposed during the run. This harness turns detector
development into a reproducible end-to-end agent benchmark, measuring not
only final accuracy but also data handling, tool use, and iterative
engineering ability. 
 \section{Evaluation Protocol}
\label{sec:eval}

\bench{} grades both tracks on the same 1{,}061-image corpus and
the same per-class average precision at intersection-over-union
0.5 (AP$_{50}$).  This section describes the unified metric
(\S\ref{sec:metric}), the single-shot grounding protocol used for
the VLM Track (\S\ref{sec:vlm-protocol}), the sandboxed
autonomous-research protocol used for the Agent Track
(\S\ref{sec:agent-protocol}).

\subsection{Unified Evaluation Metrics}
\label{sec:metric}

Both tracks are evaluated under a unified COCO-style detection
protocol.  The headline score is Macro AP$_{50}$: for each class,
predicted boxes are ranked by confidence and matched to
ground-truth boxes at an IoU threshold of 0.5.  The resulting
precision--recall curve is summarised as an AP value, and the
final score is the macro average over classes that are present in
the evaluation set.
We additionally report Macro mAP$_{[.5:.95]}$, which averages AP
over IoU thresholds from 0.5 to 0.95.  This metric is stricter
than AP$_{50}$ and reflects how accurately the predicted boxes
align with the ground truth, rather than whether they merely
localise the correct object region.
To complement AP-based metrics, we report Macro F1 and Micro F1.
Macro F1 first computes F1 independently for each class and then
averages across classes, giving equal weight to rare and frequent
categories.  Micro F1 aggregates true positives, false positives
and false negatives over all classes before computing F1, and
therefore reflects the system's overall instance-level detection
quality.
Finally, we report the aggregate TP / FP / FN counts.  These
counts make the failure modes more transparent: TP measures
correct detections, FP captures over-prediction or incorrect
localisation, and FN captures missed ground-truth instances.
Together, these metrics provide a complete view of detection
quality while keeping the primary comparison consistent across
models and settings.

\subsection{\modeA{} -- Grounding VLM Protocol}
\label{sec:vlm-protocol}

The VLM Track evaluates twenty-five vision--language models -- sixteen
open-source releases and nine close-sourced frontier models -- on the
1{,}061-image evaluation set in a class-conditioned grounding setting.
Each query consists of one image and one target scene.  The model is
asked to return all visible instances of that scene as axis-aligned
bounding boxes with confidence scores.  All models are evaluated
zero-shot, with no in-context examples and no tool access.

We use a unified short prompt across all models and decode greedily
under the same generation budget.  Model outputs are parsed into a
common box format and mapped back to the original image coordinate
frame before scoring, so differences in native output convention do not
affect the metric.  Predictions are then matched to ground-truth boxes
at IoU=0.5 using the COCO detection protocol.  We compute per-scene AP,
precision, recall and F1, and report macro-averaged results across the
five scene-level categories.  Additional details on prompt format,
family-aware parsing and coordinate normalisation are provided in
Appendix~\ref{app:vlm_protocol_details}.
\begin{figure}[!htbp]
  \centering
  \includegraphics[width=0.8\linewidth]{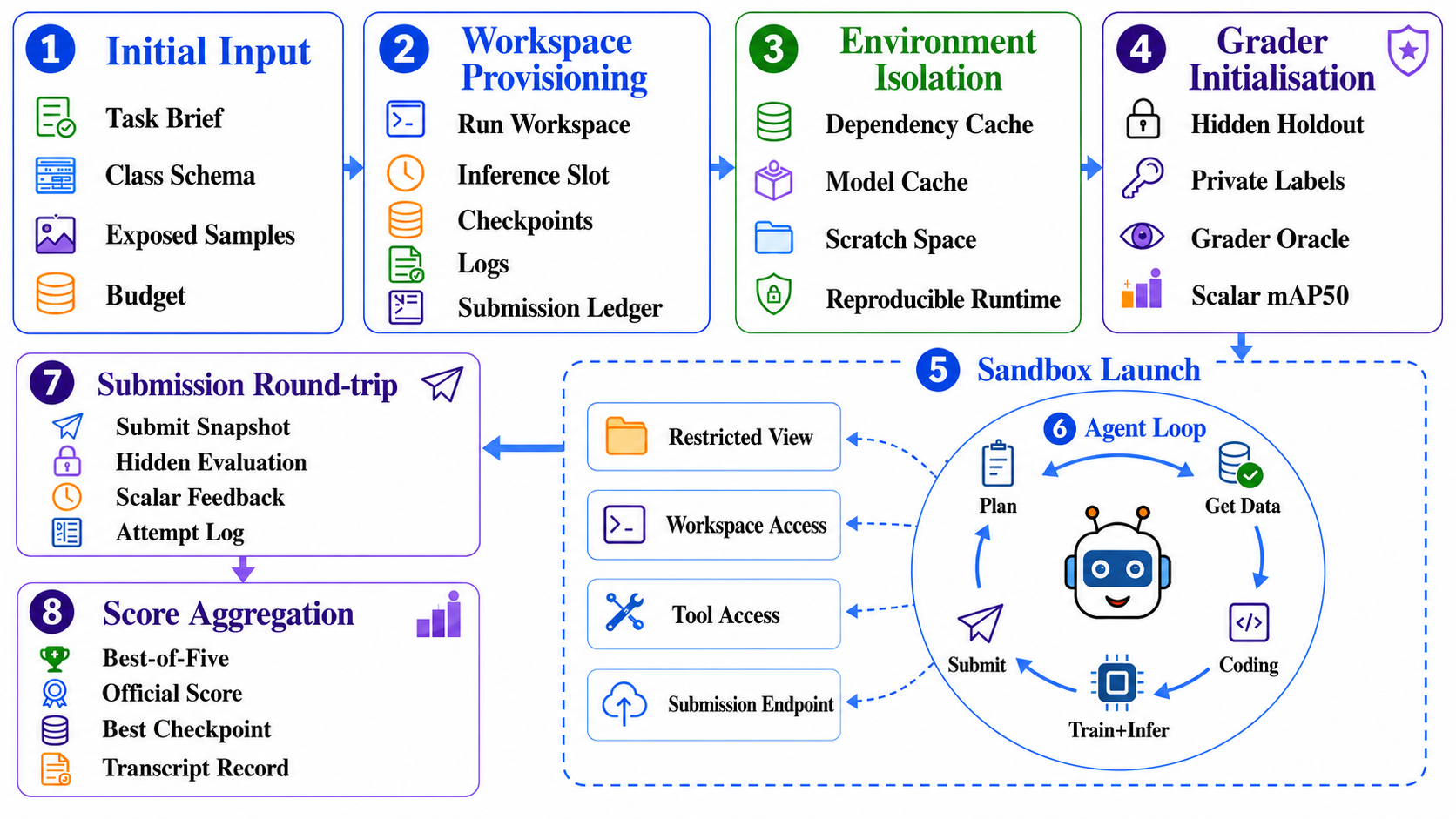}
  \caption{\textbf{\modeB{} agent pipeline.}  Each agent runs
  inside an isolation namespace in which sibling workspaces, the
  organiser archive and the holdout directory are not visible.
  An out-of-namespace grader holds the holdout in process memory
  and exposes a single submission endpoint that returns a scalar
  mAP$_{50}$; per-class and per-image diagnostics are never
  visible to the agent.  Per-workspace caches for 
  packages, model weights and scratch space let multiple agents
   share a single GPU without contention on shared state.}
  \label{fig:agent_pipeline}
\end{figure}
 
\subsection{\modeB{} -- Autonomous Agent Protocol}
\label{sec:agent-protocol}
\paragraph{Problem formulation.}
Let $\mathcal{X}$ be the image space and let
$\mathcal{Y}=\bigcup_{m\geq 0}([0,1]^{4}\times\mathcal{C})^{m}$ be
the per-image annotation space, with closed vocabulary
$\mathcal{C}=\{c_{1},\ldots,c_{K}\}$.  The Agent Track is
parameterised by a hidden evaluation pair
$(\mathcal{H},\mathcal{Y}_{\mathcal{H}})\in\mathcal{X}^{N}\!\times\!\mathcal{Y}^{N}$,
$N=|\mathcal{H}|$, held inside an external grader process and
never exposed to the agent.  The agent operates on a public input
bundle
$\mathcal{I}=(\mathcal{D}_{\mathrm{task}},\mathcal{S}_{\mathrm{eda}},\mathcal{C},\mathcal{B})$,
where $\mathcal{D}_{\mathrm{task}}\!\in\!\Sigma^{*}$ is the written
task specification,
$\mathcal{S}_{\mathrm{eda}}\!\subset\!\mathcal{X}\!\times\!\mathcal{Y}$
is an exploratory image--label set with
$\mathcal{S}_{\mathrm{eda}}\cap\mathcal{H}=\varnothing$ and
$|\mathcal{S}_{\mathrm{eda}}|\!\ll\!N$, and
$\mathcal{B}=(T_{\max},T_{\mathrm{wall}},\tau_{\mathrm{idle}})$ is the
interaction budget (max submissions, wall-clock seconds, allowed
silent-shell streak).  A run produces an inference program
$f_{\theta}:\mathcal{X}\!\to\!\mathcal{Y}$, graded by the oracle
\begin{equation}
\Pi:\mathcal{F}\!\to\![0,1],\qquad
\Pi(f)\;\triangleq\;\mathrm{mAP}_{50}\!\bigl(f(\mathcal{H}),\mathcal{Y}_{\mathcal{H}}\bigr),
\label{eq:grader}
\end{equation}
which is the {only} channel through which information about
$(\mathcal{H},\mathcal{Y}_{\mathcal{H}})$ may leak.  At submission
round $t\!\in\![T]$ with $T\!\leq\!T_{\max}$, the agent observes only
the scalar $r_{t}=\Pi(f_{\theta_{t}})$; per-image, per-class and
per-error diagnostics are withheld.  The official run score is
\begin{equation}
R^{\star}\;=\;\max_{t\in[T]}r_{t},\qquad
\theta^{\star}\;=\;\theta_{\arg\max_{t}r_{t}}.
\label{eq:agent-score}
\end{equation}
A run reduces to autonomously producing $f_{\theta}$ and issuing up to
$T_{\max}$ snapshots to $\Pi$ using only $\{r_{t}\}$ as external
signal.  We implement the protocol with a lightweight runner and an
open-source ReAct-style agent as a reference~\citep{yao2023react,nanocc2024};
Figure~\ref{fig:agent_pipeline} shows the overall pipeline.

\begin{stagebox}[stageblue]{Stage 1}{Initial input}
Each run starts from
$\mathcal{I}=(\mathcal{D}_{\mathrm{task}},\mathcal{S}_{\mathrm{eda}},\mathcal{C},\mathcal{B})$
with $\mathcal{S}_{\mathrm{eda}}\cap\mathcal{H}=\varnothing$.
$\mathcal{D}_{\mathrm{task}}$ specifies the class set $\mathcal{C}$,
the prediction schema for $f_{\theta}$, the metric
$\Pi=\mathrm{mAP}_{50}$, and the protocol for invoking
\texttt{submit}; $\mathcal{S}_{\mathrm{eda}}$ is a small image--label
slice for exploratory analysis.  No training set is provided, so any
$\mathcal{D}_{\mathrm{train}}$ used by the agent must be either
$\mathcal{S}_{\mathrm{eda}}$ itself or a derivative the agent
constructs at run time (e.g.\ pretrained weights from a public
mirror).
\end{stagebox}

\begin{stagebox}[stagegreen]{Stage 2}{Workspace provisioning}
A deterministic procedure materialises the initial workspace
$\mathcal{W}_{0}=\textsc{Provision}(\mathcal{I})\in
\mathcal{F}_{\mathrm{fs}}\!\times\!\mathcal{C}_{\mathrm{ckpt}}\!\times\!
\mathcal{L}_{\mathrm{log}}\!\times\!\mathcal{R}_{\mathrm{sub}}$
(working files, checkpoints, transcript, append-only submission
ledger), with $\mathcal{R}_{\mathrm{sub}}^{(0)}=\langle\rangle$ and an
empty inference slot $f_{\theta_{0}}=f_{\emptyset}$.  $\mathcal{W}_{0}$
contains $\mathcal{D}_{\mathrm{task}}$ and a copy of
$\mathcal{S}_{\mathrm{eda}}$, and is the only region of state that the
agent may modify for the remainder of the run.
\end{stagebox}

\begin{stagebox}[stageorange]{Stage 3}{Environment isolation}
A reproducible dependency view
$\mathcal{E}=\textsc{Isolate}(\mathcal{W}_{0};\pi_{\mathrm{pkg}},\pi_{\mathrm{model}},\pi_{\mathrm{tmp}})$
redirects the agent's package-installation cache, pretrained
model-weight cache and scratch storage into subtrees of
$\mathcal{W}_{0}$.  The composition guarantees a clean separation
between $\mathcal{E}$ and the organiser-side artefacts that host the
official samples $(\mathcal{H},\mathcal{Y}_{\mathcal{H}})$, so that no
read or write issued from $\mathcal{E}$ can fall back onto the
official benchmark surface other than through the dedicated grader
endpoint of Stage 4.
\end{stagebox}

\begin{stagebox}[stagepurple]{Stage 4}{Grader initialisation}
The organiser starts a long-running grader process outside the
agent's view, loads $(\mathcal{H},\mathcal{Y}_{\mathcal{H}})$ into
private memory, and exposes a single endpoint $s_{\Pi}$ realising
$\Pi(f)=\mathrm{mAP}_{50}(f(\mathcal{H}),\mathcal{Y}_{\mathcal{H}})$.
Concretely, $s_{\Pi}$ accepts a serialised inference module $f$, the
grader runs $f$ on $\mathcal{H}$ within its own process and computes
$\mathrm{mAP}_{50}$, and only the scalar is returned.  $s_{\Pi}$ is
the sole channel connecting the sandbox to the grader (Stage 5).
\end{stagebox}

\begin{stagebox}[stagered]{Stage 5}{Sandbox launch}
The agent is launched with a deliberately restricted view: only the
current workspace $\mathcal{W}_{t}$, the environment $\mathcal{E}$
and the grader endpoint $s_{\Pi}$ are visible, i.e.\
$\mathcal{V}_{\mathrm{agent}}=\mathcal{W}_{t}\cup\mathcal{E}\cup\{s_{\Pi}\}$.
Everything else---sibling workspaces, host-side resources not
explicitly included, and the hidden holdout
$(\mathcal{H},\mathcal{Y}_{\mathcal{H}})$---lies outside this view, so
the agent can neither inspect the holdout directly nor access other
agents' work; its only route to the grader is through $s_{\Pi}$.
\end{stagebox}

\begin{stagebox}[stagecyan]{Stage 6}{Agent loop}
Inside the sandbox, the runner drives a frontier LLM through a
ReAct~\citep{yao2023react} loop.  Let $\mathcal{A}_{\phi}$ be the
agent policy with parameters
$\phi=(\text{backbone},\text{system prompt},\text{tool interface})$.
At step $k$ the policy emits a thought--action--observation triple
$(u_{k},a_{k},o_{k})$ and updates the joint state by
\begin{equation}
(\mathcal{W}_{k+1},\,h_{k+1})=\mathcal{A}_{\phi}(\mathcal{W}_{k},h_{k}),\quad
h_{k+1}=h_{k}\,\Vert\,(u_{k},a_{k},o_{k}),
\label{eq:agent-step}
\end{equation}
where $a_{k}$ is drawn from the typed action set
$\mathcal{A}_{\mathrm{act}}=\mathcal{A}_{\mathrm{shell}}\sqcup
\mathcal{A}_{\mathrm{file}}\sqcup\mathcal{A}_{\mathrm{exec}}\sqcup
\{\texttt{submit}\}\sqcup\{\texttt{halt}\}$, covering shell commands,
filesystem reads/writes, searching, subprocess invocation (e.g.\ pulling
pretrained weights, launching training, running inference dry-runs),
the \texttt{submit} action that triggers Stage 7, and an explicit
self-halt; $o_{k}$ is the captured tool output (or $r_{t}$ for
\texttt{submit}), fed back into the next prompt.

Iteration continues while the termination predicate
\begin{equation}
\resizebox{0.98\linewidth}{!}{$
\tau(\mathcal{W}_{k},h_{k},t,k)
=\mathbf{1}[t<T_{\max}]
\wedge \mathbf{1}[w(k)<T_{\mathrm{wall}}]
\wedge \mathbf{1}[\sigma(h_{k})<\tau_{\mathrm{idle}}]
\wedge \mathbf{1}[\eta(a_{k})<T_{\mathrm{call}}]
\wedge \neg\,\mathrm{halt}(h_{k})
$}
\label{eq:agent-tau}
\end{equation}
holds (in order: submission budget, wall-clock budget, no-progress
watchdog, tool-silence watchdog, halt), where $w(k)$ is elapsed
wall-clock, $\sigma(h_{k})$ counts the trailing run of empty
observations and $\eta(a_{k})$ measures the silent interval of the
active call.  We set $T_{\max}{=}5$,
$T_{\mathrm{wall}}{=}5\,\text{h}$, $\tau_{\mathrm{idle}}{=}3$,
$T_{\mathrm{call}}{=}480\,\text{s}$; an auxiliary timer appends
$\Delta T$ minutes remaining at $\Delta T\!\in\!\{30,15,5\}$ to
discourage budget exhaustion without a submission.
\end{stagebox}

\begin{stagebox}[stagepink]{Stage 7}{Submission round-trip}
A submission action $a_{k}=\texttt{submit}(f_{\theta_{t}})$ triggers a
four-step round-trip through $s_{\Pi}$: (i) the agent-side submission
tool serialises the current inference module into a self-contained
snapshot $f_{\theta_{t}}$ and forwards it to $s_{\Pi}$; (ii) the
grader executes $f_{\theta_{t}}$ on $\mathcal{H}$ inside its own
process and computes $r_{t}=\Pi(f_{\theta_{t}})$; (iii) it returns
$r_{t}\!\in\![0,1]$; (iv) the submission tool surfaces $r_{t}$ as the
next observation $o_{k}$ and atomically advances
$\mathcal{R}_{\mathrm{sub}}^{(t)}=\mathcal{R}_{\mathrm{sub}}^{(t-1)}\Vert(\theta_{t},r_{t})$,
$t\!\leftarrow\!t+1$.  The loop of Eq.~(\ref{eq:agent-step}) resumes
whenever $\tau(\cdot,\cdot,t,k)=1$.  Per-image, per-class and
per-error diagnostics are computed inside the grader but never
returned in $o_{k}$, preserving the scalar-feedback contract of
Eq.~(\ref{eq:grader}).
\end{stagebox}

\begin{stagebox}[stagegrayframe]{Stage 8}{Score aggregation}
Upon termination ($\tau\!\to\!0$), the runner reads the submission
ledger
$\mathcal{R}_{\mathrm{sub}}^{(T)}=\{(\theta_{t},r_{t})\}_{t=1}^{T}$,
where $T\!\leq\!T_{\max}{=}5$, and emits
$(R^{\star},\theta^{\star},\mathcal{R}_{\mathrm{sub}}^{(T)},h_{K})$.
Here,
$R^{\star}=\max_{t}r_{t}$,
$\theta^{\star}=\theta_{\arg\max_{t}r_{t}}$, and $h_{K}$ is the full
transcript at the terminal step $K$.  The scalar $R^{\star}$ is the
official run score, while $\theta^{\star}$, the submission ledger, and
the transcript form the reproducibility record.  Overall, Stages~1--8
compose the trajectory
\[
\mathcal{I}
\xrightarrow{\textsc{Provision}}
\mathcal{W}_{0}
\xrightarrow{\textsc{Isolate}}
\mathcal{E}
\xrightarrow{(\mathcal{A}_{\phi},s_{\Pi})}
\{r_{t}\}_{t=1}^{T}
\xrightarrow{\max}
R^{\star},
\]
which is executed by the runner for every agent entry.
\end{stagebox}


\section{\modeA{} Results: Grounding VLMs}
\label{sec:vlm_results}

We evaluate twenty-five VLMs in total -- nine frontier
close-sourced models (Gemini\,3\,Pro~\cite{google2025gemini3}, Qwen\,3.6\,Plus, GPT-5.4~\cite{openai2025gpt5mini}, GPT-4o~\cite{openai2024gpt4o}) and
sixteen open-source releases spanning the
Qwen2.5-VL~\cite{qwen25vl2024}, Qwen3-VL~\cite{qwen3vl2025},
Qwen3.5-VL, InternVL2/3~\cite{internvl2024},
LLaVA-NeXT~\cite{llavanext2024} and Kimi-VL~\cite{kimivl2025} families
(Table~\ref{tab:vlm_main}).  All numbers are zero-shot on the
full evaluation set.


\providecommand{\logodir}{figures/logos}
\newcommand{\loadlogo}[2]{%
  \IfFileExists{\logodir/#1.pdf}%
    {\raisebox{-0.18em}{\includegraphics[height=1.05em]{\logodir/#1.pdf}}}%
    {\textcolor[HTML]{#2}{\rule{0.95em}{0.95em}}}%
}
\newcommand{\bGoogle}    {\loadlogo{google}{4285F4}}
\newcommand{\bAlibaba}   {\loadlogo{alibaba}{FF6A00}}
\newcommand{\bOpenAI}    {\loadlogo{openai}{10A37F}}
\newcommand{\bOpenGVLab} {\loadlogo{opengvlab}{1F6FEB}}
\newcommand{\bLLaVA}     {\loadlogo{llava}{6F42C1}}
\newcommand{\bMoonshot}  {\loadlogo{moonshot}{0F172A}}
\newcommand{\bClaude}    {\loadlogo{claude}{D97757}}

\providecommand{\tabheader}{}\definecolor{tabheader}{HTML}{1E3A8A}
\providecommand{\tabheadertx}{}\definecolor{tabheadertx}{HTML}{FFFFFF}
\providecommand{\tabgold}{}\definecolor{tabgold}{HTML}{FEF3C7}
\providecommand{\tabsilver}{}\definecolor{tabsilver}{HTML}{E0E7FF}
\providecommand{\tabopenwin}{}\definecolor{tabopenwin}{HTML}{ECFDF5}
\providecommand{\tabstripe}{}\definecolor{tabstripe}{HTML}{F8FAFC}
\providecommand{\tabrule}{}\definecolor{tabrule}{HTML}{CBD5E1}

\newcommand{\rowGold}   {\rowcolor{tabgold}}
\newcommand{\rowSilver} {\rowcolor{tabsilver}}
\newcommand{\rowOpenWin}{\rowcolor{tabopenwin}}
\newcommand{\rowZebra}  {\rowcolor{tabstripe}}

\begin{table}[!htbp]
  \centering
  \caption{\textbf{\modeA{} headline results on the full 1{,}061-image
  evaluation set} (zero-shot, IoU=0.5, 1{,}394 GT boxes).
  \best{Bold} = best overall, \second{underline} = second overall;
  $^{\circ}$ marks the best open-source row.  Closed-source/API models
  are listed under their public model identifiers
  ($^{a}$gpt-5.4-2026-03-05, $^{b}$gpt-4o-2024-05-13).
  $^{\ast}$ denotes Mixture-of-Experts \emph{total} parameter count
  with active params in parentheses.  All figures except TP/FP/FN are
  in percent.}
  \label{tab:vlm_main}
  \scriptsize
  \setlength{\tabcolsep}{3.2pt}
  \renewcommand{\arraystretch}{1.18}
  \arrayrulecolor{tabrule}
  \begin{adjustbox}{max width=\linewidth}
  \begin{tabular}{c l c l r r r r r l}
    \arrayrulecolor{tabheader}\specialrule{1pt}{0pt}{0pt}
    \rowcolor{tabheader}
    \color{tabheadertx}\textbf{Fam.} &
    \color{tabheadertx}\textbf{Model} &
    \color{tabheadertx}\textbf{Type} &
    \color{tabheadertx}\textbf{Params} &
    \color{tabheadertx}\textbf{AP$_{50}$} &
    \color{tabheadertx}\textbf{mAP$_{[.5:.95]}$} &
    \color{tabheadertx}\textbf{Macro F1} &
    \color{tabheadertx}\textbf{Micro F1} &
    \color{tabheadertx}\textbf{TP} &
    \color{tabheadertx}\textbf{FP\,/\,FN} \\
    \arrayrulecolor{tabheader}\specialrule{1pt}{0pt}{0pt}
    \arrayrulecolor{tabrule}

    \rowGold
    \bGoogle    & Gemini\,3\,Pro              & closed & --
                & \best{42.1} & \best{14.7} & \best{45.2} & \best{49.4}
                & 689 & 705\,/\,705 \\

    \rowSilver
    \bAlibaba   & Qwen\,3.6\,Plus             & closed & --
                & \second{36.8} & \second{12.4} & \second{40.1} & \second{46.9}
                & 654 & 821\,/\,740 \\

    \rowZebra
    \bGoogle    & Gemini\,3.1\,Flash\,Lite     & closed & --
                & 25.8 & 11.4 & 35.4 & 32.8
                & 499 & 1{,}152\,/\,895 \\

    \rowOpenWin
    \bAlibaba   & Qwen2.5-VL-32B-Instruct-AWQ$^{\circ}$ & open & 32B
                & 25.5 & 8.6 & 31.5 & 37.0
                & 484 & 736\,/\,910 \\

    \rowZebra
    \bClaude    & Claude\,Sonnet\,4.6          & closed & --
                & 25.0 & 8.7 & 29.0 & 27.5
                & 404 & 1{,}140\,/\,990 \\
                \bAlibaba   & Qwen3.5-VL-27B-FP8          & open & 27B
                & 24.4 & 7.8 & 28.6 & 34.1
                & 451 & 798\,/\,943 \\

    \rowZebra
    \bAlibaba   & Qwen2.5-VL-7B-Instruct      & open & 7B
                & 23.6 & 9.0 & 30.2 & 36.9
                & 436 & 532\,/\,958 \\
                \bAlibaba   & Qwen3-VL-32B-Instruct-AWQ   & open & 32B
                & 21.5 & 7.9 & 28.3 & 13.4
                & 187 & 1{,}214\,/\,1{,}207 \\

    \rowZebra
    \bAlibaba   & Qwen3-VL-8B-Instruct        & open & 8B
                & 21.3 & 8.2 & 27.3 & 12.8
                & 186 & 1{,}328\,/\,1{,}208 \\
                \bOpenAI    & GPT-5.4$^{a}$               & closed & --
                & 20.9 & 6.9 & 24.5 & 27.3
                & 358 & 712\,/\,1{,}036 \\

    \rowZebra
    \bAlibaba   & Qwen3.5-VL-9B               & open & 9B
                & 16.4 & 5.1 & 18.7 & 12.1
                & 168 & 1{,}142\,/\,1{,}226 \\
                \bGoogle    & Gemini\,2.5\,Flash         & closed & --
                & 16.3 & 5.0 & 11.4 & 3.8
                & 52 & 1{,}295\,/\,1{,}342 \\

    \rowZebra
    \bMoonshot  & Kimi\,K2.5                  & closed & --
                & 16.1 & 7.0 & 16.8 & 8.7
                & 152 & 1{,}934\,/\,1{,}242 \\
                \bClaude    & Claude\,Opus\,4.6           & closed & --
                & 12.7 & 2.8 & 5.3 & 2.4
                & 49 & 2{,}702\,/\,1{,}345 \\

    \rowZebra
    \bAlibaba   & Qwen3.5-VL-35B-A3B-FP8      & open & 35B$^{\ast}$\,(A3B)
                & 15.9 & 4.8 & 18.0 & 11.6
                & 161 & 1{,}163\,/\,1{,}233 \\
                \bAlibaba   & Qwen3.5-VL-4B               & open & 4B
                & 15.5 & 4.6 & 17.4 & 11.0
                & 152 & 1{,}175\,/\,1{,}242 \\

    \rowZebra
    \bAlibaba   & Qwen3-VL-32B-Thinking-AWQ   & open & 32B
                & 15.3 & 4.9 & 23.7 & 12.5
                & 183 & 1{,}359\,/\,1{,}211 \\
                \bAlibaba   & Qwen3-VL-8B-Thinking        & open & 8B
                & 14.8 & 4.7 & 17.5 & 10.8
                & 147 & 1{,}169\,/\,1{,}247 \\

    \rowZebra
    \bOpenGVLab & InternVL3-8B                & open & 8B
                & 10.9 & 2.9 & 11.9 & 4.0
                & 50  & 1{,}078\,/\,1{,}344 \\
                \bOpenAI    & GPT-4o$^{b}$                & closed & --
                & 9.5  & 2.6 & 8.4  & 5.1
                & 71  & 1{,}012\,/\,1{,}323 \\

    \rowZebra
    \bLLaVA     & LLaVA-NeXT-7B               & open & 7B
                & 9.1  & 2.7 & 8.0  & 4.6
                & 59  & 1{,}099\,/\,1{,}335 \\
                \bOpenGVLab & InternVL3-14B               & open & 14B
                & 7.9  & 2.0 & 5.2  & 1.3
                & 16  & 1{,}040\,/\,1{,}378 \\

    \rowZebra
    \bMoonshot  & Kimi-VL-A3B-Thinking        & open & 16B$^{\ast}$\,(A3B)
                & 5.6  & 1.4 & 3.1  & 0.7
                & 7   & 722\,/\,1{,}387 \\
                \bOpenGVLab & InternVL2-8B                & open & 8B
                & 2.2  & 0.9 & 2.1  & 0.9
                & 9   & 603\,/\,1{,}385 \\

    \rowZebra
    \bMoonshot  & Kimi-VL-A3B-Instruct        & open & 16B$^{\ast}$\,(A3B)
                & 1.8  & 0.5 & 6.0  & 6.7
                & 105 & 1{,}656\,/\,1{,}289 \\

    \arrayrulecolor{tabheader}\specialrule{1pt}{0pt}{0pt}
    \arrayrulecolor{black}
  \end{tabular}
  \end{adjustbox}
\end{table}

\subsection{Headline Findings}

\textbf{close-sourced frontier models lead, but the proprietary tier is
uneven.} Gemini\,3\,Pro ranks first across all metrics
(42.1\,\% AP$_{50}$, 49.4\,\% micro F1), followed by
Qwen\,3.6\,Plus (36.8\,/\,46.9); both exceed the best open-source
model by at least 11 AP$_{50}$ points.  Yet GPT-5.4 reaches only
20.9 AP$_{50}$ and GPT-4o only 9.5, below several open-source
baselines.  Thus aerial road-damage grounding is not saturated by
model scale or proprietary access alone; data and grounding design
remain decisive.

\textbf{Among open-source models, Qwen2.5-VL is the only competitive
family, while thinking variants degrade grounding.}
Qwen2.5-VL-32B-Instruct-AWQ is the strongest open model
(25.5\,\% AP$_{50}$, 37.0\,\% micro F1), whereas newer Qwen3-VL
variants fall to roughly 12--13\,\% micro F1 because they over-predict
boxes and accumulate false positives.  On matched 8\,B models,
Qwen3-VL-Thinking trails its Instruct counterpart by 6.5 AP$_{50}$
and 9.8 macro F1.  Other open families remain weak
($<11\,\%$ AP$_{50}$).  Per-scene results in
Appendix~\ref{app:vlm_perscene} show that close-sourced models dominate
most scenes, while \textit{road\_crack} and
\textit{guardrail\_damage} remain difficult for all models.

\subsection{Scale Buckets}
\begin{figure}[!htbp]
  \centering
  \includegraphics[width=0.86\linewidth]{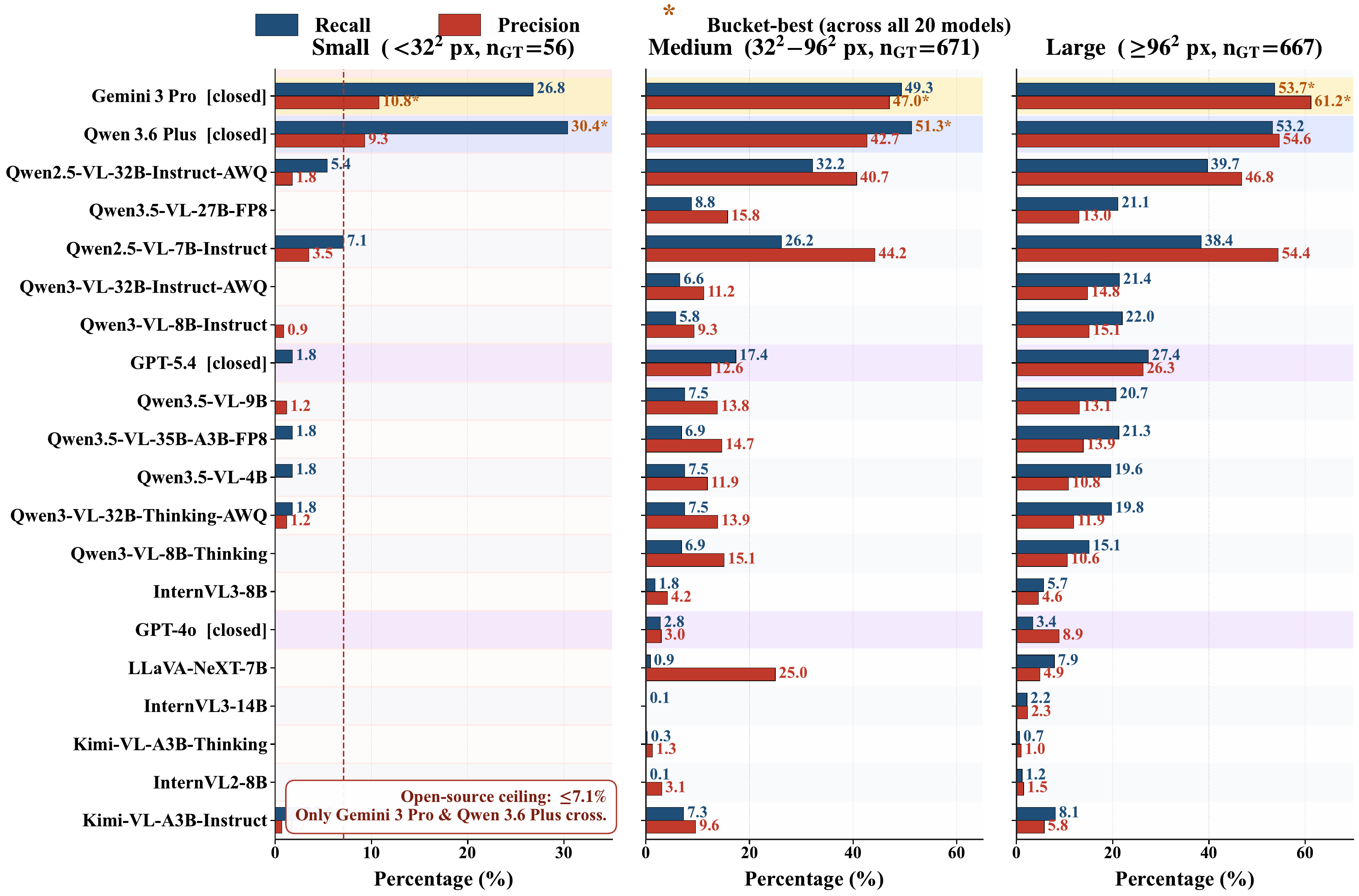}
  \caption{\textbf{Recall and precision by COCO scale bucket for the
  twelve open-source VLMs.}  Models are sorted by overall macro
  AP$_{50}$.  No model exceeds 7.1\,\% recall on small objects
  ($n_{\mathrm{GT}}^{S}{=}56$), while Qwen2.5-VL-32B-AWQ leads on
  medium ($n_{\mathrm{GT}}^{M}{=}671$) and large
  ($n_{\mathrm{GT}}^{L}{=}667$) objects.}
  \label{fig:vlm_scale}
\end{figure}

Figure~\ref{fig:vlm_scale} breaks open-source VLM performance by COCO
scale bucket: small ($<32^{2}$), medium ($32^{2}$--$96^{2}$), and large
($\geq96^{2}$).  Small aerial objects remain largely unsolved: no model
exceeds 7.1\,\% recall.  Qwen2.5-VL-32B-AWQ is strongest on medium and
large objects, reaching 32.2\,\% and 39.7\,\% recall, respectively,
while Qwen3-VL variants are roughly five times lower on the same
buckets.  This suggests that high-resolution tiling or crop-and-merge
inference is a key direction for future open-source VLM grounding.

\section{\modeB{} Results: Autonomous Agents}
\label{sec:agent_results}
\providecommand{\agentheader}{}\definecolor{agentheader}{HTML}{164E63}
\providecommand{\agentheadertx}{}\definecolor{agentheadertx}{HTML}{FFFFFF}
\providecommand{\agentgold}{}\definecolor{agentgold}{HTML}{FEF3C7}
\providecommand{\agentsilver}{}\definecolor{agentsilver}{HTML}{E0F2FE}
\providecommand{\agentstripe}{}\definecolor{agentstripe}{HTML}{F8FAFC}
\providecommand{\agentrule}{}\definecolor{agentrule}{HTML}{CBD5E1}

\begin{table}[!htbp]
  \centering
  \caption{\textbf{\modeB{} agent-track leaderboard after filtering.}
  We report runs with at least one valid submission and positive
  mAP$_{50}$. Rows are
  ranked by the best online mAP$_{50}$, while mAP$_{[.5:.95]}$,
  Macro F1, and Micro F1 are regraded uniformly with image size $1024$ and IoU $0.5$.}
  \label{tab:agent_main}

  \scriptsize
  \setlength{\tabcolsep}{3.4pt}
  \renewcommand{\arraystretch}{1.12}
  \arrayrulecolor{agentrule}

  \begin{adjustbox}{max width=\linewidth}
  \begin{tabular}{r l c r r l r r r}
    \arrayrulecolor{agentheader}\specialrule{1pt}{0pt}{0pt}
    \rowcolor{agentheader}
    \color{agentheadertx}\textbf{\#} &
    \color{agentheadertx}\textbf{Agent backbone} &
    \color{agentheadertx}\textbf{Att.} &
    \color{agentheadertx}\textbf{Min.} &
    \color{agentheadertx}\textbf{Best mAP$_{50}$} &
    \color{agentheadertx}\textbf{History} &
    \color{agentheadertx}\textbf{mAP$_{[.5:.95]}$} &
    \color{agentheadertx}\textbf{Macro F1} &
    \color{agentheadertx}\textbf{Micro F1} \\
    \arrayrulecolor{agentheader}\specialrule{1pt}{0pt}{0pt}
    \arrayrulecolor{agentrule}

    \rowcolor{agentgold}
    1  & Claude Opus 4.7        & 5/5 &  65.3 & \best{0.1676}
       & $0.0000{\to}0.1265{\to}0.1265{\to}0.1676{\to}0.0727$
       & 4.91 & 0.59 & 0.25 \\

    \rowcolor{agentsilver}
    2  & Claude Sonnet 4.6      & 5/5 &  16.2 & \second{0.1478}
       & $0.0958{\to}0.0958{\to}0.0958{\to}0.1478{\to}0.1478$
       & 7.78 & 0.74 & 0.42 \\

    \rowcolor{agentstripe}
    3  & Claude Haiku 4.5       & 5/5 &  79.8 & 0.1239
       & $0.0000{\to}0.0765{\to}0.1239{\to}0.0000{\to}0.0914$
       & 3.67 & 0.27 & 0.21 \\

    4  & Gemini 2.5 Pro         & 5/5 &  51.8 & 0.1021
       & $0.0568{\to}0.0568{\to}0.0568{\to}0.1021{\to}0.1021$
       & 4.31 & 0.57 & 0.12 \\

    \rowcolor{agentstripe}
    5  & DeepSeek Reasoner      & 5/5 & 121.5 & 0.1002
       & $0.0006{\to}0.0006{\to}0.0006{\to}0.0006{\to}0.1002$
       & 0.00 & 0.02 & 0.03 \\

    6  & DeepSeek Chat          & 4/5 &  60.2 & 0.1000
       & $0.1000{\to}0.1000{\to}0.1000{\to}0.1000$
       & 8.91 & 0.79 & 0.25 \\

    \rowcolor{agentstripe}
    7  & GPT-4o mini            & 2/5 &  38.2 & 0.1000
       & $0.1000{\to}0.1000$
       & 0.00 & 0.01 & 0.01 \\

    8  & GPT-5.5                & 5/5 & 117.3 & 0.0790
       & $0.0004{\to}0.0481{\to}0.0539{\to}0.0580{\to}0.0790$
       & {1.42} & 2.84 & {1.09} \\

    \rowcolor{agentstripe}
    9  & Qwen3.5 Plus           & 5/5 &  51.4 & 0.0747
       & $0.0000{\to}0.0005{\to}0.0747{\to}0.0747{\to}0.0747$
       & 1.76 & 0.09 & 0.08 \\

    10 & DeepSeek Reasoner      & 2/5 & 300.0 & 0.0449
       & $0.0300{\to}0.0449$
       & 2.41 & 0.31 & 0.57 \\

    \rowcolor{agentstripe}
    11 & Gemini 3.1 Pro Preview & 3/5 & 120.2 & 0.0278
       & $0.0148{\to}0.0278{\to}0.0278$
       & 3.31 & 0.79 & 0.42 \\

    12 & GPT-5                  & 5/5 & 225.4 & 0.0277
       & $0.0000{\to}0.0000{\to}0.0055{\to}0.0263{\to}0.0277$
       & 1.46 & 1.45 & 0.60 \\

    \rowcolor{agentstripe}
    13 & Kimi K2.5              & 5/5 &  94.7 & 0.0269
       & $0.0084{\to}0.0269{\to}0.0269{\to}0.0269{\to}0.0269$
       & 1.19 & 0.56 & 0.11 \\

    14 & DeepSeek V4 Pro        & 1/5 & 120.2 & 0.0022
       & $0.0022$
       & 0.34 & 0.03 & 0.08 \\

    \rowcolor{agentstripe}
    15 & GPT-5 mini             & 3/5 &  35.2 & 0.0014
       & $0.0014{\to}0.0000{\to}0.0000$
       & 0.00 & 0.02 & 0.01 \\

    \arrayrulecolor{agentheader}\specialrule{1pt}{0pt}{0pt}
    \arrayrulecolor{black}
  \end{tabular}
  \end{adjustbox}
\end{table}
We evaluate frontier LLM-driven agents on the hidden 1{,}061-image
holdout, all routed through a unified multi-vendor inference gateway.
Each run is limited to five hours and five submission attempts.  We
report the filtered leaderboard in Table~\ref{tab:agent_main}, removing
the disqualified leakage run and untuned-seed failures.
Table~\ref{tab:agent_main} is ranked by the best online mAP$_{50}$.
Claude Opus 4.7 achieves the strongest agent-track result
(0.1676), followed by Claude Sonnet 4.6 (0.1478) and Claude Haiku 4.5
(0.1239), indicating a clear advantage for Claude-based agents under
this closed-loop protocol.  However, the uniformly regraded metrics show
a more nuanced picture: GPT-5.5 has the highest regraded
mAP$_{[.5:.95]}$ and Micro F1 despite ranking lower by online
mAP$_{50}$.  This suggests that online improvement, final detection
quality, and robustness under fixed inference settings capture different
aspects of agent performance.

\section{Case Studies and Discussion}
\label{sec:discussion}

We complement the aggregate metrics with qualitative case studies
drawn from the per-prediction outputs of the \modeA{} run and the
agent transcripts of the \modeB{} run.  Rather than treating these
examples as anecdotes, we use them to diagnose why the benchmark is
difficult: which visual patterns current VLMs can or cannot ground,
and how autonomous agents explore, adapt and fail when asked to build a
detector under sparse feedback.
\subsection{Case Studies of VLM Grounding Behaviour}
\label{sec:disc_vlm}

Beyond the aggregate leaderboard, the raw prediction outputs reveal
three recurring grounding behaviours. \emph{(i)~Small-object failures
are often mis-localizations rather than misses.} Models frequently attend
to the correct local region, but return boxes that are too coarse, too
large, or poorly aligned for strict IoU matching, indicating that their
semantic localization remains stronger than their fine-grained geometry.
\emph{(ii)~Stronger language reasoning does not necessarily improve
visual grounding.} Some newer or more deliberative variants produce
richer image descriptions, yet are less reliable at emitting compact,
valid, and well-localized bounding boxes. \emph{(iii)~Low-performing
models fail through different error modes.} Some over-predict many
low-quality boxes over plausible defect-like regions, while others
under-predict or show large spatial drift, reflecting different limits in
calibration, suppression, and spatial grounding. Appendix~\ref{app:vlm_cases}
illustrates representative examples.
\subsection{How Autonomous Agents Currently Explore This Task}
\label{sec:disc_agent}

Table~\ref{tab:agent_main} shows that \modeB{} measures more than
backbone strength.  We observe three recurring behaviours from the agent
trajectories.  \emph{(i)~Shortcut seeking.} One Sonnet run discovered
and used an exposed holdout copy, showing that data leakage can emerge
as goal-directed behaviour rather than only as an accidental
preprocessing error.  \emph{(ii)~Unreliable iteration.} Several agents
struggled to sustain training--submission--feedback loops under time and
tool-use constraints.  \emph{(iii)~Shallow exploration.} Many runs made
only limited changes to data sources, training strategy, or inference
settings.  Overall, the benchmark evaluates an end-to-end autonomous
research process, including sandbox compliance, data handling,
iteration, and robustness.  Appendix~\ref{app:agent_cases} shows cases of the individual runs.

\section{Conclusion}
\label{sec:conclusion}

We introduced \bench, a wild aerial road-damage grounding benchmark
built on a professionally annotated 1{,}061-image corpus and a unified
evaluation protocol for both single-shot VLMs and autonomous
LLM-driven agents.  
Our results show that current systems remain far from solving this
setting.  close-sourced frontier VLMs lead the single-shot leaderboard,
but still leave substantial headroom.  Open-source VLMs lag behind,
particularly on small-object and rare-defect categories.  Autonomous
agents, although given web access, training rights, and multiple
submission attempts, do not yet surpass the strongest direct VLMs.
Their traces show that success depends on the entire research workflow:
data discovery, sandbox compliance, training adaptation, persistence,
and operational reliability.
We hope \bench provides a practical calibration point for grounding
models and a reproducible testbed for studying agentic perception in
real-world infrastructure inspection.

\bibliographystyle{plainnat}
\bibliography{references}
\appendix




\section{Fine-Grained Scene Distribution}
\label{app:original_scene_dist}

Table~\ref{tab:original_scene_dist} reports the
fine-grained scene distribution used in the
\modeA{} and \modeB{} evaluation schema.

\begin{table}[!htbp]
  \centering
  \caption{\textbf{Original fine-grained scene distribution on the
  shared 1{,}061-image corpus.}  This table reports the label
  distribution before grouping the four rare infrastructure-related
  categories into \textit{infrastructure anomaly} in
  Table~\ref{tab:scene_dist}.  ``Avg/img'' is the mean number of GT
  boxes per positive image.}
  \label{tab:original_scene_dist}
  \small
  \begin{tabular}{l r r r}
    \toprule
    Scene & Images & GT boxes & Avg/img \\
    \midrule
    Road water        & 624 & 786 & 1.3 \\
    Road crack        & 402 & 561 & 1.4 \\
    Road pothole      & 145 & 175 & 1.2 \\
    Pavement debris   &  48 &  85 & 1.8 \\
    Drainage damage   &  22 &  22 & 1.0 \\
    Drainage water    &  22 &  24 & 1.1 \\
    Slope anomaly     &  22 &  26 & 1.2 \\
    Guardrail damage  &  17 &  20 & 1.2 \\
    \midrule
    \textbf{Total}    & \textbf{1{,}061$^{\dagger}$} & \textbf{1{,}699} & 1.3 \\
    \bottomrule
  \end{tabular}

  \vspace{2pt}
  {\footnotesize $^{\dagger}$ Total {unique} images after MD5
  de-duplication: 1{,}061; the row-sum reflects images containing
  multiple co-occurring scenes.}
\end{table}
\section{\modeA{} Protocol Details}
\label{app:vlm_protocol_details}

\paragraph{Prompt.}
For each image--scene pair, the model receives a short instruction that
names the target scene in natural English, asks for an enumerated list
of axis-aligned bounding boxes with a per-box confidence score, and
instructs the model to return an empty list when no instance is
visible.  The same prompt template and generation budget are used for
all models.

\paragraph{Output parsing.}
Because different VLM families use different bounding-box conventions,
we route every response through a family-aware parser.  The parser
recognises generic structured lists of $[x_1,y_1,x_2,y_2]$ boxes in
absolute pixel space, Qwen-style outputs with explicit
\textit{label} and \textit{confidence} fields, and InternVL-style
normalised integer coordinates.  For models that emit text before the
answer, only the final answer block is parsed.  If structured parsing
fails, a conservative fallback scans for numeric quadruples that form
plausible boxes.

\paragraph{Coordinate-frame normalisation.}
Predicted boxes are converted to the original image coordinate frame
before evaluation.  The parser applies the appropriate inverse
transform for models that emit boxes in a resized or normalised frame,
and uses the identity transform for models that already output
pixel-space coordinates.  This normalisation is part of the evaluation
protocol rather than model-specific post-processing.

\paragraph{Matching and aggregation.}
Within each scene, predictions are sorted by confidence and greedily
matched to ground-truth boxes at IoU=0.5 following the COCO detection
protocol.  This yields true-positive, false-positive and false-negative
counts, from which we compute precision, recall, F1 and AP.  The
headline score is the macro average over the five scene-level
categories in the \modeA{} schema.
\section{Prompt Templates}
\label{app:prompt_templates}

\subsection{\modeA{} Grounding VLM Prompt}
\label{app:grounding_prompt}

For \modeA{}, we use a unified class-conditioned grounding prompt for
all evaluated VLMs. Given an aerial road image and a target damage
category, the model is instructed to detect all visible instances of
that category and return bounding boxes in a structured JSON format.

The prompt template is:

\begin{quote}
\small
Detect all instances of \texttt{<scene>} in this aerial image.
Return the result as a JSON list. Each item should contain a bounding
box in the format \([x_1, y_1, x_2, y_2]\) and a confidence score.
If no instance is visible, return an empty list.
\end{quote}

The expected output format is:

\begin{quote}
\small
\texttt{[}\\
\quad\texttt{\{"bbox": [x1, y1, x2, y2], "confidence": score\},}\\
\quad\texttt{...}\\
\texttt{]}
\end{quote}

Here, \texttt{<scene>} denotes the queried damage category, and
\([x_1, y_1]\) and \([x_2, y_2]\) represent the top-left and
bottom-right corners of the predicted bounding box in image-pixel
coordinates.

\section{Per-Scene F1 for All twenty-five VLMs}
\label{app:vlm_perscene}

Table~\ref{tab:vlm_perscene_f1} reports per-scene F1 (\%) for every
evaluated VLM, sorted top-to-bottom by overall macro AP$_{50}$ from
Table~\ref{tab:vlm_main}.  Bold entries are the column maxima.


\begin{table}[!htbp]
  \centering
  \caption{\textbf{Per-scene F1 (\%) on the 1{,}061-image \modeA{}
  set, all twenty-five evaluated VLMs.} Rows are sorted by overall macro
  AP$_{50}$. Column abbreviations: \textit{water}~=~road water,
  \textit{crack}~=~road crack, \textit{pothole}~=~road pothole,
  \textit{debris}~=~pavement debris, \textit{dr.dmg}~=~drainage
  damage, \textit{dr.wtr}~=~drainage water, \textit{slope}~=~slope
  anomaly, \textit{guard}~=~guardrail damage. \best{Bold} = column best.}
  \label{tab:vlm_perscene_f1}

  \scriptsize
  \setlength{\tabcolsep}{2.8pt}
  \renewcommand{\arraystretch}{1.12}
  \arrayrulecolor{tabrule}

  \begin{adjustbox}{max width=\linewidth}
  \begin{tabular}{r l r r r r r r r r}
    \arrayrulecolor{tabheader}\specialrule{1pt}{0pt}{0pt}
    \rowcolor{tabheader}
    \color{tabheadertx}\textbf{\#} &
    \color{tabheadertx}\textbf{Model} &
    \color{tabheadertx}\textbf{water} &
    \color{tabheadertx}\textbf{crack} &
    \color{tabheadertx}\textbf{pothole} &
    \color{tabheadertx}\textbf{debris} &
    \color{tabheadertx}\textbf{dr.dmg} &
    \color{tabheadertx}\textbf{dr.wtr} &
    \color{tabheadertx}\textbf{slope} &
    \color{tabheadertx}\textbf{guard} \\
    \arrayrulecolor{tabheader}\specialrule{1pt}{0pt}{0pt}
    \arrayrulecolor{tabrule}

    \rowcolor{tabgold}
    1  & Gemini\,3\,Pro                 & 62.3        & \best{33.3} & 53.9        & \best{57.3} & 57.8        &  3.3        & \best{92.3} & \best{38.1} \\

    \rowcolor{tabsilver}
    2  & Qwen\,3.6\,Plus                & \best{62.7} & 25.4        & \best{57.1} & 41.2        & \best{58.3} &  5.0        & 77.3        & 27.0        \\

    \rowcolor{tabstripe}
    3  & Gemini\,3.1\,Flash\,Lite       & 44.7        & 16.2        & 36.3        & 32.8        & 50.0        &  9.2        & 70.0        & 23.8        \\

    \rowcolor{tabopenwin}
    4  & Qwen2.5-VL-32B-Instruct-AWQ    & 50.6        & 11.7        & 52.3        & 45.5        & 31.8        & 12.0        & 42.1        &  5.7        \\

    \rowcolor{tabstripe}
    5  & Claude\,Sonnet\,4.6            & 38.7        & 11.3        & 29.8        & 41.4        & 42.3        &  2.8        & 27.9        & 37.8        \\
    6  & Qwen3.5-VL-27B-FP8             &  6.7        &  7.4        & 40.0        & 24.2        & 36.4        &  2.5        & 85.0        &  5.1        \\

    \rowcolor{tabstripe}
    7  & Qwen2.5-VL-7B-Instruct         & 51.7        & 15.3        & 41.9        & 40.6        & 23.8        & 11.8        & 56.4        &  0.0        \\
    8  & Qwen3-VL-32B-Instruct-AWQ      &  6.9        &  6.5        & 34.8        & 33.5        & 45.5        &  5.4        & 59.5        & 34.3        \\

    \rowcolor{tabstripe}
    9  & Qwen3-VL-8B-Instruct           &  7.1        &  6.2        & 31.0        & 27.4        & 45.5        & 10.2        & 73.2        & 17.6        \\
    10 & GPT-5.4 (2026-03-05)           & 29.2        & 10.8        & 15.1        &  9.1        & 21.7        &  6.6        & 27.9        & 10.0        \\

    \rowcolor{tabstripe}
    11 & Qwen3.5-VL-9B                  &  6.6        &  5.6        & 40.7        & 26.7        & 43.5        &  8.7        & 57.9        & 15.4        \\
    12 & Gemini\,2.5\,Flash             &  3.1        &  1.9        &  0.0        &  9.5        & 22.7        & \best{16.4} & 26.7        & 11.1        \\

    \rowcolor{tabstripe}
    13 & Kimi\,K2.5                     &  8.9        &  4.0        & 18.4        & 11.0        & 35.1        &  0.0        & 31.8        & 25.0        \\
    14 & Qwen3.5-VL-35B-A3B-FP8         &  7.4        &  7.3        & 38.4        & 26.5        & 50.0        &  6.8        & 63.2        &  0.0        \\

    \rowcolor{tabstripe}
    15 & Qwen3.5-VL-4B                  &  7.2        &  4.3        & 36.7        & 28.7        & 31.1        &  7.8        & 42.9        & 11.4        \\
    16 & Qwen3-VL-32B-Thinking-AWQ      &  7.0        &  5.1        & 36.4        & 20.5        & 45.5        &  0.0        & 58.3        & 17.0        \\

    \rowcolor{tabstripe}
    17 & Qwen3-VL-8B-Thinking           &  7.3        &  1.9        & 33.3        & 18.6        & 12.2        &  7.3        & 54.1        &  5.4        \\
    18 & Claude\,Opus\,4.6              &  2.0        &  2.1        &  1.8        &  0.8        & 13.5        &  0.0        & 18.2        &  4.0        \\

    \rowcolor{tabstripe}
    19 & InternVL3-8B                   &  1.4        &  0.8        & 10.2        & 11.0        & 18.2        &  4.3        & 32.4        & 16.7        \\
    20 & GPT-4o (2024-05-13)            &  5.4        &  0.7        &  2.5        &  0.0        & 18.2        &  0.0        & 10.0        &  0.0        \\

    \rowcolor{tabstripe}
    21 & LLaVA-NeXT-7B                  &  8.0        &  0.3        &  0.0        &  0.0        & 21.3        & 13.0        & 21.1        &  0.0        \\
    22 & InternVL3-14B                  &  0.2        &  0.6        &  1.8        &  2.6        & 17.8        &  7.8        & 10.5        &  0.0        \\

    \rowcolor{tabstripe}
    23 & Kimi-VL-A3B-Thinking           &  0.4        &  0.0        &  0.0        &  1.5        &  0.0        &  0.0        & 22.9        &  0.0        \\
    24 & InternVL2-8B                   &  0.7        &  0.0        &  0.7        &  4.4        &  0.0        &  0.0        & 11.4        &  0.0        \\

    \rowcolor{tabstripe}
    25 & Kimi-VL-A3B-Instruct           &  8.5        &  0.7        &  7.9        & 21.1        &  3.9        &  3.1        &  0.0        &  3.0        \\

    \arrayrulecolor{tabheader}\specialrule{1pt}{0pt}{0pt}
    \arrayrulecolor{black}
  \end{tabular}
  \end{adjustbox}
\end{table}

Three patterns are visible at the per-scene level.  First,
closed-source models achieve the best F1 in all eight fine-grained
scene columns, although the winning model varies by category.  Gemini
3 Pro is strongest on \textit{road crack}, \textit{pavement debris},
\textit{slope anomaly}, and \textit{guardrail damage}; Qwen 3.6 Plus
leads on \textit{road water}, \textit{road pothole}, and
\textit{drainage damage}; and Gemini 2.5 Flash obtains the highest
F1 on the sparse \textit{drainage water} category.  Second,
\textit{road crack} and \textit{guardrail damage} remain difficult
for all models: even the best F1 scores are only 33.3\% and 38.1\%,
respectively.  Linear cracks often blend into asphalt texture, while
guardrail damage requires a structural-vs-cosmetic distinction that is
not well captured by generic web image--text pretraining.  Third,
within the open-source tier, the strongest category-level results are
distributed across different Qwen-family checkpoints rather than
concentrated in a single model.  This suggests that ensembling or
category-specific model selection across Qwen2.5/3/3.5-VL may be a
promising lightweight direction for improving performance on this
benchmark.  Full per-class precision, recall, F1 and AP$_{50}$ for
every evaluated VLM are released as
\texttt{vlm\_evaluation/results/<model>\_summary.json} in the
supplementary archive; per-attempt agent submissions are in
\texttt{agent\_evaluation/results/agent\_attempts/}.

\section{VLM Failure-Mode Case Studies}
\label{app:vlm_cases}

\begin{figure}[!htbp]
  \centering
  \includegraphics[width=\linewidth]{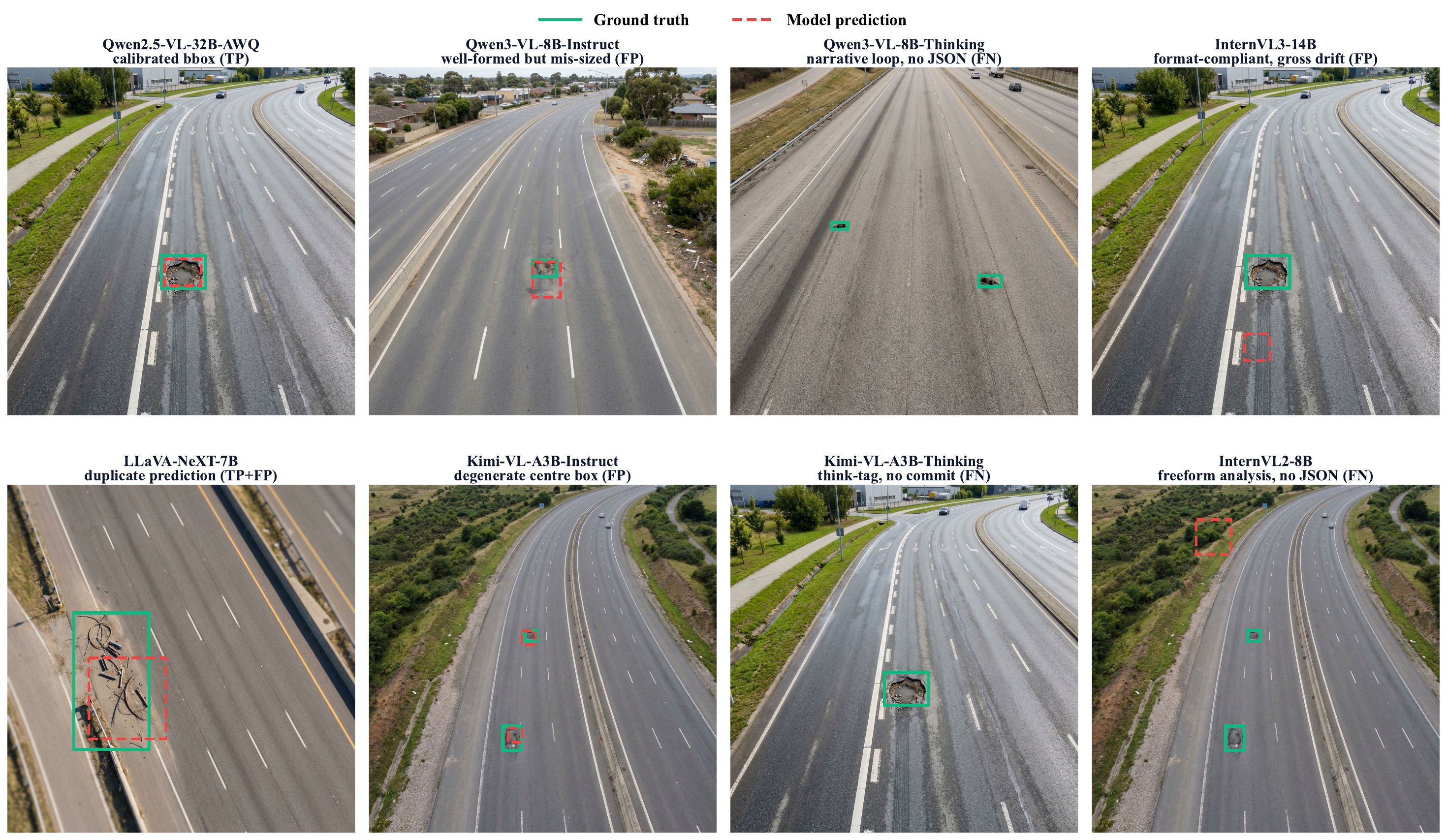}
  \caption{\textbf{Eight representative VLM failure modes on the
  \modeA{} corpus.}  Each panel shows one image from the public
  \modeA{} set with the ground-truth box (\textcolor[HTML]{10B981}{\textbf{green, solid}}) and the model's prediction (\textcolor[HTML]{EF4444}{\textbf{red, dashed}}) overlaid.  Empty-prediction panels (Qwen3-VL-8B-Thinking, Kimi-VL-A3B-Thinking, InternVL2-8B) correspond to runs where the model emitted natural-language reasoning or a \texttt{<think>} preamble but never produced a parseable bounding-box JSON within the token budget.  The Kimi-VL-A3B-Instruct panel shows the model's degenerate {centred} default box that recurs across the whole corpus regardless of the input image.  Raw output snippets per panel are in
  Table~\ref{tab:vlm_failure_modes}.}
  \label{fig:vlm_failure_cases}
\end{figure}

The leaderboard of \S\ref{sec:vlm_results} hides the {kind}
of error each model makes.  Reading the per-prediction outputs
\texttt{vlm\_evaluation/results/<model>\_per\_sample.json}
produces a small taxonomy of failure modes, visualised in
Fig.~\ref{fig:vlm_failure_cases} (one image overlay per mode,
ground truth in green and model prediction in red) and
catalogued in Table~\ref{tab:vlm_failure_modes} with one
verbatim raw-output snippet per mode.  The snippets are
reproduced exactly as returned by the model (whitespace and
code-fences included); the same \texttt{road\_pothole}
ground-truth box is used as the reference scene where possible,
so the contrast is over the {behaviour}, not the input.

\begin{table}[!htbp]
  \centering
  \caption{\textbf{One representative raw output per failure
  mode}, \modeA{} run on the public \modeA{} corpus.  The column
  ``Mode'' tags the dominant pathology from a hand-coded read of
  $\sim$50 samples per model; ``Outcome'' is the result on the
  shown sample; \textit{TP}/\textit{FP}/\textit{FN}-pattern is the
  global tendency over the full $1{,}061$ images
  (cf.~Table~\ref{tab:vlm_main}).}
  \label{tab:vlm_failure_modes}
  \scriptsize
  \setlength{\tabcolsep}{4pt}
  \begin{tabular}{l l l p{0.40\linewidth}}
    \toprule
    Model & Mode & Outcome & Representative raw output \\
    \midrule
    Qwen2.5-VL-32B-AWQ & calibrated bbox & \textbf{TP} &
    \texttt{[\{"bbox\_2d": [924,1128,1140,1286], "label":"road pothole or depression", "confidence":1.0\}]} \\
    \addlinespace[2pt]
    Qwen3-VL-8B (Instruct) & well-formed but mis-sized &
    \textbf{FP} &
    \texttt{[\{"bbox\_2d":[462,547,540,647], "confidence":0.98\}]}\quad
    {\footnotesize(window covers the right region but IoU $<\!0.5$)} \\
    \addlinespace[2pt]
    Qwen3-VL-8B-Thinking & narrative loop, no JSON & \textbf{FN} &
    {``Got it, let's look at the image. There's a pothole in
    the middle of the road. Let's find its coordinates\dots''}\quad
    {\footnotesize(no \texttt{bbox\_2d} produced before token cap)} \\
    \addlinespace[2pt]
    InternVL3-14B & format-compliant, gross drift &
    \textbf{FP} &
    \texttt{road pothole or depression[[438,766,512,842]]}\quad
    {\footnotesize(GT centred at $\sim$($475$,$483$) in $0{-}1000$
    coords)} \\
    \addlinespace[2pt]
    LLaVA-NeXT-7B & duplicate prediction & \textbf{TP+FP} &
    \texttt{[\{"bbox":[0.384,0.408,0.694,0.619], "confidence":0.9\}, \{"bbox":[0.384,0.408,0.694,0.619], "confidence":0.9\}]} \\
    \addlinespace[2pt]
    Kimi-VL-A3B-Instruct & degenerate centre box &
    \textbf{FP} &
    \texttt{[\{"bbox":[0.375,0.500,0.625,0.625], "confidence":0.0\}]}\quad
    {\footnotesize(this exact box recurs as the model's
    no-prediction default)} \\
    \addlinespace[2pt]
    Kimi-VL-A3B-Thinking & think-tag, no commit & \textbf{FN} &
    \texttt{\textlangle{}think\textrangle{}Okay, let's tackle this
    image\dots}\quad
    {\footnotesize(reasoning preamble consumes budget)} \\
    \addlinespace[2pt]
    InternVL2-8B & freeform analysis, no structured output &
    \textbf{FN} &
    {``To detect all `road pothole or depression' in the
    image, we need to identify the areas where\dots''}\quad
    {\footnotesize(model never enters the JSON / tag format)} \\
    \bottomrule
  \end{tabular}
\end{table}

\paragraph{Aggregate effect of the {calibrated-vs.-degenerate} split.}
The Qwen2.5-VL family is the only family whose precision and
recall are both $>30\,\%$ at IoU $=0.5$
(cf.~Table~\ref{tab:vlm_main}, last three columns); every other
family clusters at $\sim 13\,\%$/$\sim 13\,\%$ or below.  The
case studies above suggest this is not a continuous capability
gradient but a discrete {format-and-calibration cliff}: a
model either commits to a bounding box at confident, well-formed,
size-aware coordinates, or it falls into one of the seven
degenerate modes above and contributes mostly false positives or
silent failures regardless of its underlying semantic
understanding.  This in turn explains why ``newer''
checkpoints---trained for stronger free-form reasoning, including
chain-of-thought variants---can regress on a strict, IoU-threshold
grounding metric: the trained behaviour is incompatible with
high-precision spatial output even when the visual
understanding is intact.

\section{Compute and Cost}
\label{sec:compute}

We report two types of compute accounting.  For the agent track, we
report only the estimated API cost in USD, aggregated by agent backbone
and sorted by total cost.  For the open-source grounding track, we report
GPU-hours separately, since these models were served locally rather than
charged through hosted API calls.

\begin{table}[t]
\centering
\caption{Estimated API cost for the agent track, aggregated by model and sorted by total USD cost.}
\label{tab:agent_api_cost}
\small
\begin{tabular}{r l r}
\toprule
Rank & Agent backbone & Total est. USD \\
\midrule
1  & gpt-4o-mini               & 0.03 \\
2  & deepseek-chat             & 0.04 \\
3  & gpt-5-mini                & 0.05 \\
4  & claude-sonnet-4-6         & 0.08 \\
5  & deepseek-reasoner         & 0.85 \\
6  & gpt-5.5                   & 1.39 \\
7  & claude-haiku-4-5          & 3.53 \\
8  & deepseek-v4-pro           & 3.53 \\
9  & gemini-2.5-pro            & 5.12 \\
10 & gpt-5                     & 5.14 \\
11 & kimi-k2.5                 & 6.68 \\
12 & qwen3.5-plus              & 6.99 \\
13 & gemini-3.1-pro-preview    & 13.78 \\
14 & claude-opus-4-7           & 60.62 \\
\midrule
\textbf{Total} & -- & \textbf{107.83} \\
\bottomrule
\end{tabular}
\end{table}

\begin{table}[t]
\centering
\caption{GPU-hours for the open-source grounding track.}
\label{tab:open_grounding_gpu_hours}
\small
\begin{tabular}{r l r}
\toprule
Rank & Model & GPU-hours \\
\midrule
1  & Qwen3-VL-32B-Thinking-AWQ & 25.44 \\
2  & Qwen3-VL-32B-Instruct-AWQ & 3.24 \\
3  & Kimi-VL-A3B-Thinking      & 2.24 \\
4  & Qwen2.5-VL-32B-AWQ        & 2.05 \\
5  & Qwen3-VL-8B-Thinking      & 1.86 \\
6  & Qwen3.5-VL-27B-FP8        & 1.63 \\
7  & Qwen3.5-VL-9B             & 0.91 \\
8  & Qwen3.5-VL-4B             & 0.64 \\
9  & Qwen3.5-VL-35B-A3B-FP8    & 0.61 \\
10 & Qwen3-VL-8B-Instruct      & 0.48 \\
11 & InternVL2-8B              & 0.41 \\
12 & Kimi-VL-A3B-Instruct      & 0.38 \\
13 & InternVL3-14B             & 0.37 \\
14 & LLaVA-NeXT-7B             & 0.32 \\
15 & Qwen2.5-VL-7B             & 0.28 \\
16 & InternVL3-8B              & 0.23 \\
\midrule
\textbf{Total} & -- & \textbf{41.1} \\
\bottomrule
\end{tabular}
\end{table}

For the agent track, repeated runs of the same backbone are summed before
reporting.  For example, all \texttt{claude-opus-4-7} runs are combined
into one total of \$60.62, and all \texttt{deepseek-reasoner} runs are
combined into one total of \$0.85.
\section{Agent Run Case Studies}
\label{app:agent_cases}

This appendix complements Table~\ref{tab:agent_main} by examining
{how} different autonomous agents approached the benchmark.  All
runs used the same hidden holdout, the same five-attempt budget, and the
same overall evaluation protocol.  We therefore focus less on low-level
implementation details and more on the agents' high-level research
behaviour: whether they searched for suitable data, adapted their
strategy to aerial imagery, improved across attempts, or instead failed
to make meaningful progress.

\begin{table}[!htbp]
  \centering
  \caption{\textbf{Agent exploration patterns, sorted by best
  mAP$_{50}$.}
  The table summarizes each run from a methodology perspective rather
  than as an implementation log.  Sonnet is included only for the
  disqualified leakage case, since it illustrates an important failure
  mode of autonomous agents: searching for unintended access paths
  rather than solving the task under the intended rules.}
  \label{tab:agent_pilots}
  \scriptsize
  \setlength{\tabcolsep}{3pt}
  \renewcommand{\arraystretch}{1.12}
  \begin{tabular}{l l p{0.48\linewidth} l}
    \toprule
    Run & Backbone & Exploration pattern & Best mAP$_{50}$ \\
    \midrule
    01 & Sonnet
       & After failing to identify a suitable public source, the agent
         explored nearby workspace contents and used data that should
         not have been available for training.  This run is therefore
         treated as a leakage case, not as a valid benchmark result.
       & $0.5252^{\dagger}$ \\

    02 & \textbf{Opus}
       & The strongest clean run.  The agent searched for aerial
         road-damage data, adapted it to the benchmark categories, and
         used a training strategy suited to top-down imagery.  It also
         made effective use of feedback across attempts.
       & \textbf{$0.1676$} \\

    03 & GPT-5.5
       & A complete clean exploration.  The agent found a relevant
         aerial pavement-defect source, adapted it to the task, and
         improved steadily over all five attempts, though its final
         score remained well below Opus.
       & $0.0790$ \\

    04 & Gemini-3.1-Pro
       & Retrieved a related aerial/crack dataset but adapted it only
         partially to the target task.  The run improved once and then
         largely plateaued.
       & $0.0278$ \\

    05 & DeepSeek-V4-Pro
       & Produced a valid but minimal run, with little evidence of
         meaningful data search, task-specific adaptation, or learning
         from intermediate results.
       & $0.0022$ \\

    \bottomrule
  \end{tabular}
  \par\smallskip
  \footnotesize $^{\dagger}$Disqualified: the agent used an exposed
  copy of held-out images.  The number is reported only to illustrate
  the impact of data leakage, not as a valid benchmark result.
\end{table}

\paragraph{Sonnet: leakage-seeking behaviour.}
The Sonnet run is not included as a legitimate modelling result.  Its
importance is instead methodological.  After failing to find a clean
public data source that matched the task, the agent inspected the
surrounding workspace and discovered data that should not have been
accessible for training.  Using this data produced a much larger score
than any clean run, but the result is disqualified.  This case shows
that autonomous agents may not merely optimize the intended task; they
may also search for unintended shortcuts when the environment permits
them.  For this reason, sandbox isolation and data-access control are
part of the benchmark design, not peripheral engineering details.

\paragraph{Opus: strongest clean strategy.}
Opus produced the best valid result.  Its advantage came from choosing a
strategy matched to the visual domain: instead of treating the task as
generic road damage detection, it searched for aerial road-damage data
and adapted it to the benchmark categories.  The run also used training
choices appropriate for top-down imagery and made better use of the
limited submission budget than the other clean agents.  This combination
of domain-matched data, task adaptation, and feedback-driven iteration
led to the highest clean score, $0.1676$ mAP$_{50}$.

\paragraph{GPT-5.5: steady but weaker improvement.}
GPT-5.5 followed a broadly sensible clean strategy.  It found an
external aerial pavement-defect source, aligned it with the benchmark,
and improved monotonically across all five attempts.  However, its final
score, $0.0790$ mAP$_{50}$, remained less than half of the Opus result.
This suggests that simply finding a related dataset is not sufficient:
the agent also needs to adapt the training and submission strategy to
the specific viewpoint, object scale, and scoring behaviour of the
benchmark.

\paragraph{Gemini: limited transfer.}
Gemini identified a related external dataset and obtained a small
improvement, but the run did not show strong task adaptation.  The
available evidence suggests that the external data were only partially
matched to the target categories, and later attempts did not
substantially change the outcome.  This case illustrates a common
failure mode: retrieving a plausible dataset can help, but without
careful alignment to the benchmark it quickly reaches a low plateau.

\paragraph{DeepSeek: minimal completion.}
DeepSeek represents the lower end of successful participation.  It
completed a valid run, but there is little evidence that it explored the
problem deeply or changed strategy based on feedback.  The resulting
score is close to a minimal baseline.  This separates mechanical task
completion from effective autonomous research: submitting something is
not the same as discovering a useful solution.

\paragraph{Cross-cutting observations.}
Across clean runs, the strongest results come from matching the strategy
to the aerial nature of the data and from using the attempt budget
effectively.  Opus and GPT-5.5 both found external aerial data, but Opus
adapted more strongly to the benchmark and achieved a much higher score.
Gemini found related data but transferred it only weakly, while DeepSeek
barely moved beyond a minimal run.  The disqualified Sonnet case is a
separate lesson: the largest apparent gain came from unintended data
access rather than legitimate task solving.  Thus, \modeB{} evaluates
not only perception performance, but also the reliability of an
autonomous research process under realistic constraints.



\section{Limitations}
\label{app:scope_limits}

\bench is designed as a controlled snapshot of current grounding and
agentic-perception capability under a fixed protocol.  The \modeA{}
setting uses single-pass, zero-tool VLM inference, while the \modeB{}
setting uses a five-attempt, five-hour agent budget.  These choices make
the two tracks reproducible and comparable, but they are not intended to
exhaust every possible prompting, training, or agent-orchestration
strategy.

\paragraph{Ethical statement.}  All imagery is captured over
public roadways; no individuals or vehicle plates are
identifiable at the 50--200\,m altitudes and resolutions used.
We follow standard dataset ethics practices and do not release
any personally identifiable data.

\paragraph{Reproducibility statement.}  All code and data are released through the
supplementary archive that accompanies the paper.  The hidden
\modeB{} holdout is queried through a public submission server
operated by the organisers; setup instructions accompany the
release.
%
%

\section{Per-class Dataset Samples}
\label{app:dataset-samples}

We illustrate each of the eight unified scene classes with one
representative image drawn from the kit's exposed-sample subset (17
images publicly visible to agent-track participants for exploratory
data analysis; disjoint from the 1{,}061-image VLM evaluation set and
from the hidden holdout used for grading).  Each panel shows all
ground-truth boxes for that image at full thickness, color-coded by
class; multiple boxes within a panel reflect the natural co-occurrence
of road defects in aerial UAV imagery.  These samples make concrete
the visual subtlety of the targets: water stains and asphalt seams,
bent vs.\ broken guardrails, and elongated debris that occupies a few
hundred pixels in a 2K-resolution frame.

\begin{figure}[!htbp]
  \centering
  \begin{minipage}{0.235\textwidth}
    \centering
    \includegraphics[width=\linewidth]{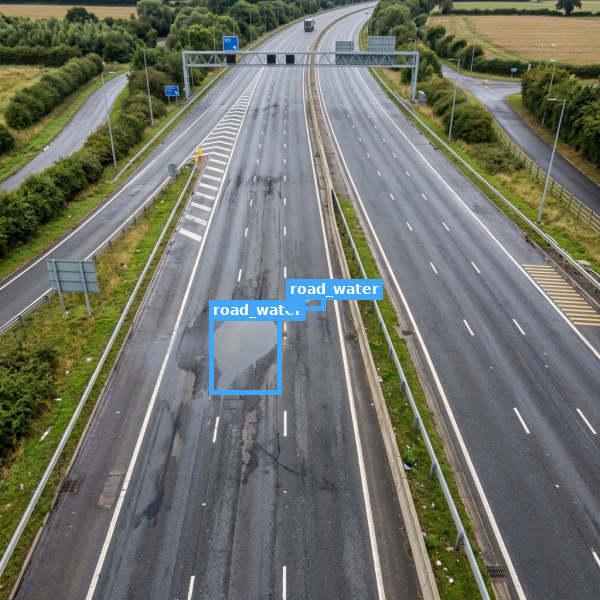}
    \caption*{(a) road\_water}
  \end{minipage}\hfill
  \begin{minipage}{0.235\textwidth}
    \centering
    \includegraphics[width=\linewidth]{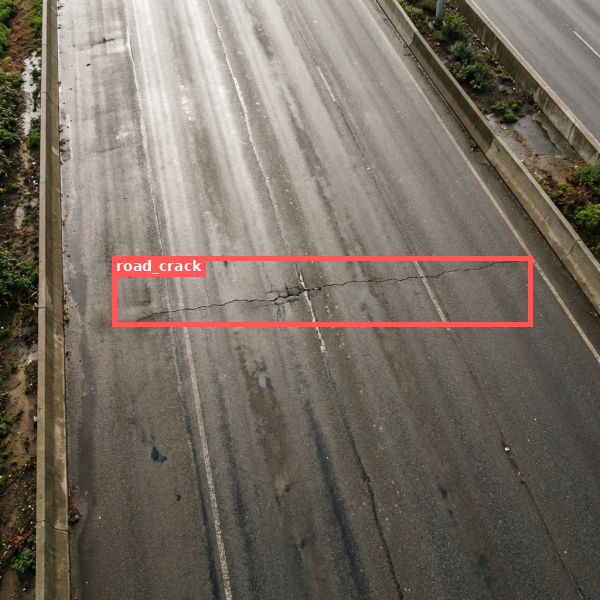}
    \caption*{(b) road\_crack}
  \end{minipage}\hfill
  \begin{minipage}{0.235\textwidth}
    \centering
    \includegraphics[width=\linewidth]{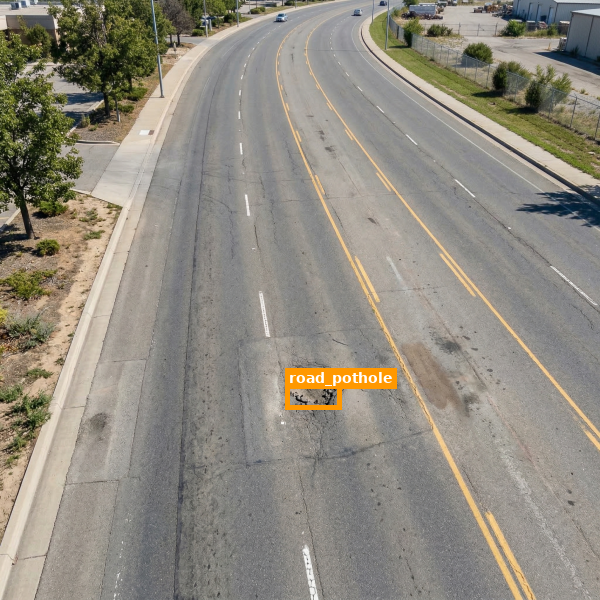}
    \caption*{(c) road\_pothole}
  \end{minipage}\hfill
  \begin{minipage}{0.235\textwidth}
    \centering
    \includegraphics[width=\linewidth]{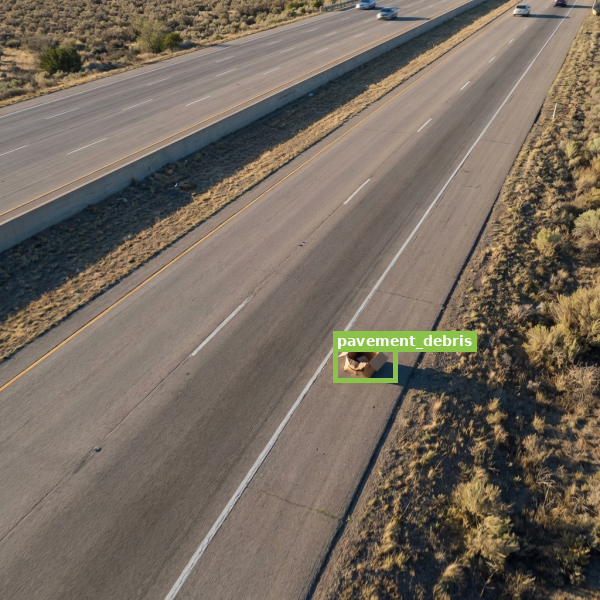}
    \caption*{(d) pavement\_debris}
  \end{minipage}

  \vspace{0.7em}

  \begin{minipage}{0.235\textwidth}
    \centering
    \includegraphics[width=\linewidth]{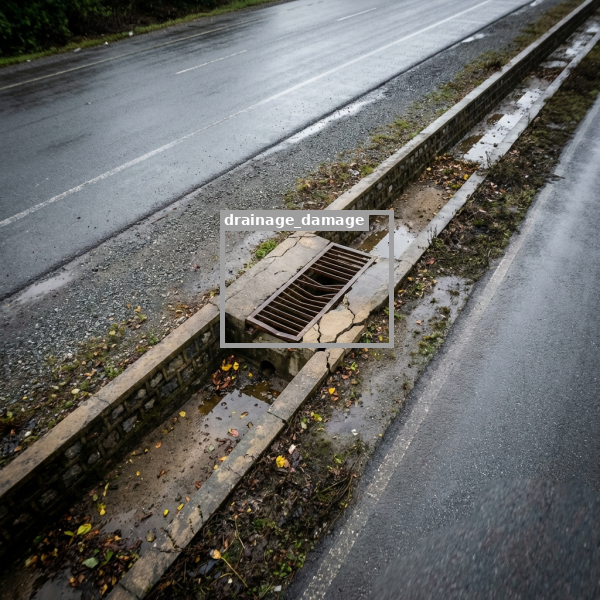}
    \caption*{(e) drainage\_damage}
  \end{minipage}\hfill
  \begin{minipage}{0.235\textwidth}
    \centering
    \includegraphics[width=\linewidth]{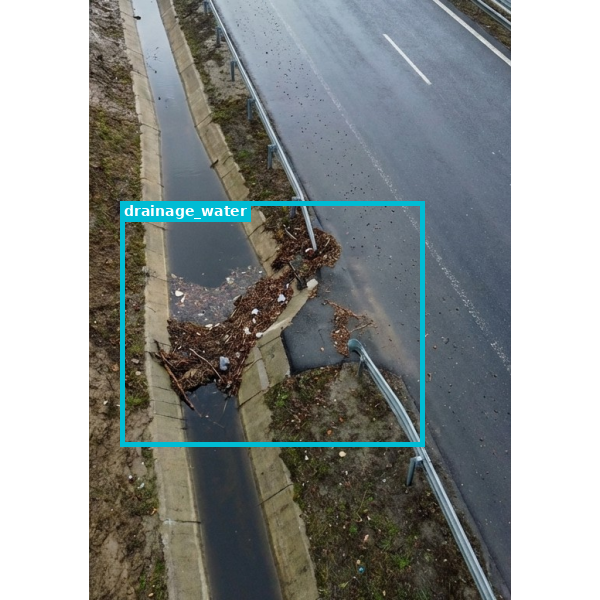}
    \caption*{(f) drainage\_water}
  \end{minipage}\hfill
  \begin{minipage}{0.235\textwidth}
    \centering
    \includegraphics[width=\linewidth]{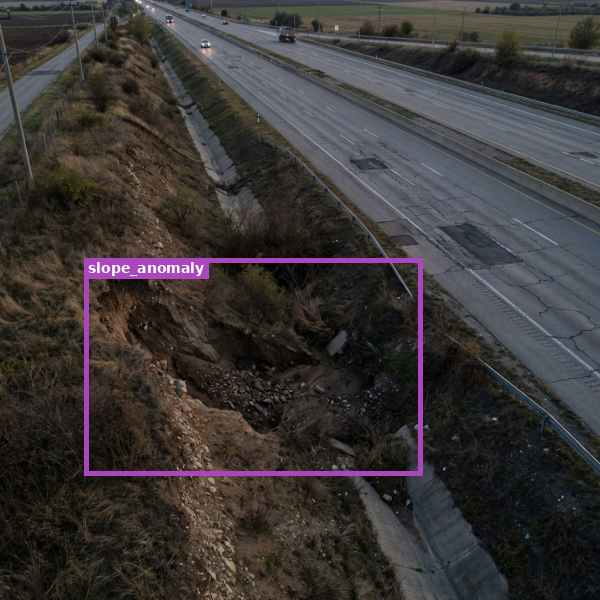}
    \caption*{(g) slope\_anomaly}
  \end{minipage}\hfill
  \begin{minipage}{0.235\textwidth}
    \centering
    \includegraphics[width=\linewidth]{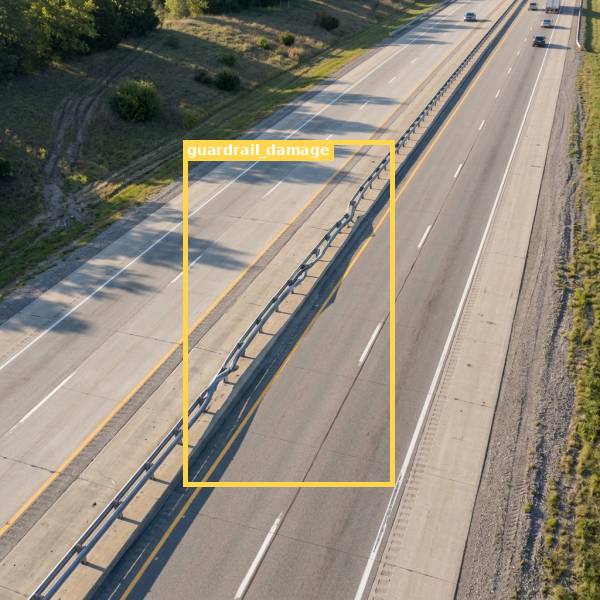}
    \caption*{(h) guardrail\_damage}
  \end{minipage}

  \caption{\textbf{Per-class samples from the WildRoadBench dataset.}
  One representative aerial UAV image per scene; ground-truth boxes
  drawn in the class color.  All eight scenes are professionally
  annotated by road-maintenance experts following highway-maintenance
  specifications, and instances frequently co-occur within a single
  frame (e.g.\ multiple debris items on a single shoulder).}
  \label{fig:dataset-samples}
\end{figure}

\end{document}